\documentclass{format/ieeetj}

\usepackage{pratik}

\usepackage[color=gray!30,textsize=tiny]{todonotes}

\usepackage{algorithm,algorithmic}

\usepackage{glossaries}
\newacronym{mav}{MAV}{Micro Aerial Vehicles}
\newacronym{uav}{UAV}{Unmanned Aerial Vehicle}
\newacronym{ovc}{OVC}{Open Vision Computer}
\newacronym{lidar}{LiDAR}{Light Detection and Ranging}
\newacronym{vio}{VIO}{visual-inertial odometry}
\newacronym{gpgpu}{GPGPU}{General-Purpose Graphics Processing Unit}
\newacronym{ugv}{UGV}{Unmanned Ground Vehicle}
\newacronym{uwb}{UWB}{Ultra Wideband}
\newacronym{svm}{SVM}{Support Vector Machine}
\newacronym{fcn}{FCN}{Fully Convolutional Network}
\newacronym{cnn}{CNN}{Convolutional Neural Network}
\newacronym{loam}{LOAM}{LiDAR Odometry and Mapping}
\newacronym{sloam}{SLOAM}{Semantic LiDAR Odometry and Mapping}
\newacronym{slam}{SLAM}{Simultaneous Localization and Mapping}
\newacronym{iot4ag}{IoT4Ag}{NSF Engineering Research Center for the Internet of Things for Precision Agriculture}
\newacronym{grasp-lab}{GRASP Lab}{the General Robotics, Automation, Sensing and Perception Laboratory}
\newacronym{jps}{JPS}{Jump Point Search}
\newacronym{ukf}{UKF}{Unscented Kalman Filter}
\newacronym{sam}{SAM}{Smoothing and Mapping}
\newacronym{icp}{ICP}{Iterative Closest Point}
\newacronym{imu}{IMU}{Inertial Measurement Unit}
\newacronym{tsdf}{TSDF}{Truncated Signed Distance Field}
\newacronym{esdf}{ESDF}{Euclidean Signed Distance Field}
\newacronym{sdf}{SDF}{Signed Distance Field}
\newacronym{rrt}{RRT}{Rapidly Exploring Random Tree}
\newacronym{fpv}{FPV}{First-person View}
\newacronym{dnn}{DNN}{Deep Neural Network}
\newacronym{igpred}{IGPred}{Information Gain Prediction}
\newacronym{csqmi}{CSQMI}{Cauchy-Schwarz Quadratic Mutual Information}
\newacronym{nbv}{NBV}{Next Best View}
\newacronym{vae}{VAE}{Variational Autoencoder}
\newacronym{tsp}{TSP}{Traveling Salesman Problem}
\newacronym{bcsm}{BCSM}{Behavior Control State Machine}
\newacronym{pca}{PCA}{Principal Component Analysis}
\newacronym{aspp}{ASPP}{Atrous Spatial Pyramid Pooling}
\newacronym{swap}{SWaP}{Size Weight and Power}
\newacronym{soi}{SoI}{Semantic Object of Interest}
\newacronym{aoi}{AoI}{Area of Interest}
\newacronym{drl}{DRL}{Deep Reinforcement Learning}
\newacronym{dl}{DL}{Deep Learning}
\newacronym{fov}{FoV}{Field of View}
\newacronym{tops}{TOPS}{Tera Operations per Second}
\newacronym{nerf}{NeRF}{Neural Radiance Field}

\usepackage{tikz}
\usetikzlibrary{shapes.geometric, arrows}

\usepackage{pgfplots}
\usepackage{arydshln}

\def\OJlogo{\vspace{-4pt}$<$Society logo(s) and publication title will appear here.$>$}
\def\seclogo{\vspace{10pt}$<$Society logo(s) and publication title will appear here.$>$}

\def\authorrefmark#1{\ensuremath{^{\textbf{#1}}}}

\title{ATLAS Navigator: Active Task-driven LAnguage-embedded Gaussian Splatting}

\receiveddate{20 February, 2026}
\reviseddate{26 June, 2026}
\accepteddate{1 September, 2026}

\author{Dexter Ong\authorrefmark{1}, Graduate Student Member, IEEE, \\
Yuezhan Tao\authorrefmark{1}, Graduate Student Member, IEEE,\\
Varun Murali\authorrefmark{2}, Member, IEEE, Igor Spasojevic \authorrefmark{3},\\
Vijay Kumar \authorrefmark{1}, Fellow, IEEE, and
Pratik Chaudhari \authorrefmark{1}, Member, IEEE}
\affil{GRASP Laboratory, University of Pennsylvania, Philadelphia, PA 19104 USA}
\affil{Texas A\&M University, College Station, TX 77840 USA}
\affil{University of California Riverside, Riverside, CA 92521 USA}

\corresp{Corresponding author: Dexter Ong (email: odexter@engineering.upenn.edu).}
\authornote{This work was supported in
part by TILOS under NSF Grant CCR- 2112665, 
in part by IoT4Ag ERC under NSF Grant EEC-1941529, 
in part by NSF under Grant CMMI-2415249, 
in part by NSF NRI/USDA under Award 2022-67021-36856,
in part by ARL DCIST under Grant CRA W911NF-17-2-0181,
in part by DSO National Laboratories,
and in part by NVIDIA.}

\begin{document}

\IEEEpeerreviewmaketitle

\begin{abstract}

We address the challenge of task-oriented
navigation in unstructured and unknown environments, where robots must incrementally build and reason on informative metric-semantic maps in real time.
Since tasks may require clarification or re-specification, it is necessary for the information in the map to be rich enough to enable generalization across a wide range of tasks.
To effectively execute tasks specified in natural language, we propose a hierarchical representation built on language-embedded Gaussian splatting that enables both sparse semantic planning that lends itself to online operation
and dense geometric representation for collision-free navigation.
We validate the effectiveness of our method through real-world robot experiments on a ground robot equipped with only vision-based sensing.
The experiments are conducted in both cluttered indoor and kilometer-scale outdoor environments, covering more than 2 \unit{\kilo\m} distance and 47,000 \unit{\square\m} of area.
We achieve a competitive ratio of about 59\% relative to the shortest possible path, demonstrating efficient exploration and semantic navigation in challenging real-world environments.
Experiment videos and more details can be found on our project page: https://ongdexter.github.io/atlasnav \\
\end{abstract}


\maketitle

\begin{figure*}
    \centering    \includegraphics[width=\linewidth,trim={0.5cm 5.0cm 1cm 5.5cm},clip]{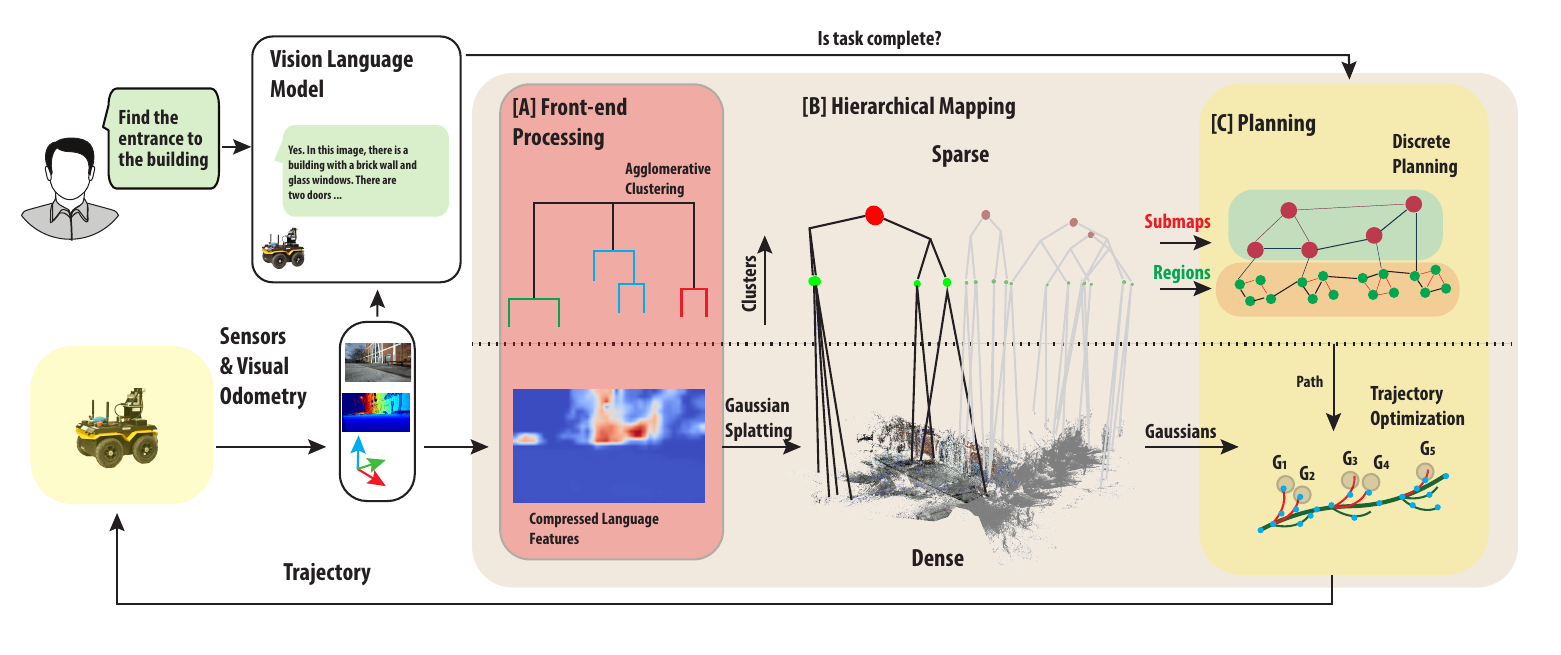}
    \caption{Our framework consists of three components.
    The front-end processing [A] extracts and compresses dense pixel-level language features from the image.
    The module also performs agglomerative clustering on the features.
    The hierarchical mapper [B] runs bottom-up, processing the RGB and depth images and odometry from the robot to build a map.
    The top level of the map contains the submaps, the middle level the regions, and the bottom level the Gaussians.
    The local map comprises the loaded submaps while the other submaps are unloaded to save memory (shown here in gray).
    The planning module [C] consists of a discrete planner that operates on the sparse map and generates a reference path, while the dense Gaussians in the local map are used to generate collision-free and dynamically feasible trajectories to be executed on the robot. A vision language model is used to determine task completion.
    }
    \label{fig:system-diagram}
\end{figure*}


\section{Introduction}
\label{sec:intro}

\IEEEPARstart{T}{he} increasing deployment of robots in real-world environments for infrastructure inspection~\cite{lattanzi2017review}, search-and-rescue~\cite{chung2023into, kruijff2012rescue} and agriculture~\cite{fountas2020agricultural} necessitates the development of systems capable of natural language interaction with humans, providing both textual and visual feedback. 
This, in turn, requires robots to autonomously perceive their surroundings, gather relevant information, and make safe and efficient decisions -- capabilities crucial for a variety of \textbf{\emph{in-the-wild tasking approaches over kilometer-scale environments with sparse semantics}}.
To enable these capabilities on board robots with compute constraints, we develop a framework to efficiently store and plan on hierarchical metric-semantic maps with visual and inertial sensors only.
An overview of our method is shown in Fig.~\ref{fig:system-diagram}.

A cornerstone of autonomous navigation is the creation of actionable maps that effectively represent the environment and support diverse navigation and task-specific operations.  
Such maps must possess several key desiderata: (1) a consistent association between semantic and geometric information derived from observations, enabling a holistic understanding of the environment; (2) an efficient storage mechanism for navigation, such as a hierarchical organization of semantic information into submaps coupled with the representation of geometric information using Gaussian distributions, providing scalability and precise spatial modeling; and (3) task-relevant identification and adaptability, achieved through scoring Gaussian components to facilitate generalization across various tasks and adaptation to new objectives.  
These properties collectively ensure that the map is not only manageable but also capable of supporting large-scale autonomous navigation to complete tasks provided in natural language.

We propose an agglomerative map representation built on 3D Gaussian Splatting (3DGS) that is consistent across geometric and semantic scales. Semantic features are extracted using a vision-language model and stored as compressed vectors alongside the Gaussian parameters (Sec.~\ref{sec:mapping}).
This unified structure enables task-driven retrieval and scalable map storage suitable for efficient large-scale navigation.
The map supports both rapid local access for on-board motion planning and retention of dense semantic context for tasks such as long-range planning and loop closure.
We integrate this representation into a two-stage framework (Sec.~\ref{sec:planning}) consisting of a high-level discrete planner and a low-level trajectory generator.
This \textbf{\emph{unified mapping and planning pipeline}}, operating from a shared vision-based sensor input, reduces computational load, avoids map inconsistencies, and enables deployment on resource-constrained robots.

The key contributions of this work are summarized as follows:
\begin{itemize}
    \item \textbf{Memory-efficient online language-embedded Gaussian splatting:}  
    We introduce a method for Gaussian splatting that incorporates language embeddings efficiently to optimize memory usage. In this work, we embed compressed language feature vectors for efficient online updates, ensuring that the map maintains a compact and scalable representation while preserving the necessary geometric and semantic fidelity.

    \item \textbf{Submapping with sparse semantic hierarchical clusters and dense geometry:}  
    To enable large-scale 3DGS mapping on a SWaP-constrained platform, we propose a submapping framework that dynamically loads submaps onto the GPU. To support efficient querying and planning on this representation, each submap is structured into sparse semantic clusters that represent regions or objects and dense geometric representations for navigation. This design leverages a tectonic structure, which updates submap anchor poses after loop closure. As a result, the map is compatible with external odometry sources and pre-built maps, enabling consistent updates and large-scale navigation.

    \item \textbf{Large-scale autonomous task-driven navigation with reusable maps:}  
    Our approach supports task-driven navigation by dynamically identifying and retrieving map regions relevant to specific objectives that are specified in natural language. The system is also compatible with real-time, interactive task re-specification, allowing for adaptive exploration and navigation in large-scale environments.
    
\end{itemize}

We demonstrate these contributions through extensive ablations that show the efficiency and semantic retrieval capability of our method (Sec.~\ref{sec:seg-experiments}), and real-world demonstrations of the full framework through large-scale vision-based autonomy on a mobile robot (Sec.~\ref{sec:robot-exp}).

\section{Related Work}
\label{sec:related_work}

Our proposed work lies at the intersection of active perception, efficient language-embedded Gaussian splatting, hierarchical metric-semantic mapping and planning, and task-driven navigation.
We propose a compact hierarchical language-embedded map representation built upon Gaussian splatting that can enable task-driven autonomous navigation while incrementally exploring and building a map of the environment.
We also provide in Table~\ref{table:related_comparison} a comparison of ATLAS against similar mapping and semantic navigation frameworks.

\begin{table*}[t]
    \centering
    \footnotesize
    \setlength{\tabcolsep}{6pt}
    \renewcommand{\arraystretch}{1.15}
    \begin{tabular}{p{2.5cm}|p{5cm}|p{2.7cm}|p{2.0cm}|p{1cm}|p{1cm}}
    \toprule
    \textbf{System} & \textbf{Semantics} & \textbf{Submapping} & \textbf{Scale} & \textbf{Vision-only} & \textbf{Real-time} \\
    \midrule
    ConceptFusion~\cite{jatavallabhula2023conceptfusion} & Coarse point-cloud features with sparse region/object features from image segments & None & Indoor scale & \ding{56} & \ding{56} \\
    ConceptGraphs~\cite{conceptgraphs} & Sparse object-centric scene graph from segment-derived object point clouds & None & Indoor scale & \ding{56} & \ding{56} \\
    VLFM~\cite{yokoyama2024vlfm} & Coarse, task-specific and frontier-level scores from 2D relevancy grid & None & Indoor scale & \ding{52} & \ding{52} \\
    Clio~\cite{maggio2024clio} & Sparse scene graph with task-specific object and region nodes & None & Indoor scale & \ding{52} & \ding{52} \\
    HOV-SG~\cite{werby2024hierarchical} & Point-wise and segment features with floor/room/object hierarchy & None & Building scale & \ding{56} & \ding{56} \\
    \textbf{ATLAS (ours)} & \textbf{Dense language-embedded Gaussians and sparse region/submap hierarchy} & \textbf{Dynamic load/unload and update} & \textbf{Indoor and kilometer-scale outdoor} & \ding{52} & \ding{52} \\
    \bottomrule
    \end{tabular}
    \caption{Positioning of ATLAS relative to related open-vocabulary semantic mapping and navigation systems. Several related approaches use image-segmentation priors to construct object- or segment-level semantic primitives, whereas ATLAS stores compressed language features directly in the dense Gaussian representation and constructs its hierarchy from those features. This sparse hierarchical structure and submapping enables online task-driven navigation in large-scale and unstructured environments.}
    \label{table:related_comparison}
\end{table*}

\subsection{Active Perception}
Active perception methods allow the robot to actively select actions and viewpoints that maximize the information gain relevant to a given task.
This problem has been widely studied in volumetric representations such as voxel maps or Signed Distance Field (SDF) maps.
In~\cite{charrow2015information, saulnier2020information, asgharivaskasi2023semantic, bai2016bayesianexp}, the information gain is evaluated as the mutual information between the current map and expected map given future observations.
To accelerate the computation, information gain can be approximated with cell counting-based methods~\cite{LukasIG, bircher2016receding, Papachristos17UncertaintyIG, alexis2020MP,  schmid2021unified, yuezhantao2023seer, tao-24-3dactivemsslam}.
The information-driven approach has also been widely adopted for autonomous exploration with learned representations such as Neural Radiance Fields (NeRFs) or neural SDFs. 
Several approaches~\cite{pan2022activenerf} estimate information gain in NeRFs by selecting future observations from a training dataset or by sampling viewpoints in the radiance fields~\cite{Lee2022nerf3drecon, zhan2022activermap, ran2023neurar, he2023active, he2024active}.
In neural SDFs, ~\cite{yan2023active} evaluates information gain as the variance of the model induced by parameter perturbations.
The approach in ~\cite{feng2024naruto} quantifies information gain by learning the reconstruction uncertainty with an MLP.
Fisher Information has been used as the proxy for evaluating mutual information for 3D Gaussian splatting~\cite{jiang2023fisherrfactiveviewselection}.
Combining a voxel map with 3DGS, in~\cite{jin2024gsplanner}, the information gain is evaluated as the weighted sum of unobserved volume between rays in the occupancy map, where the weights are determined by the transmittance of the Gaussians.
The method has been further extended in~\cite{xu2024hgs}, by incorporating Fisher Information as part of the information gain of a candidate viewpoint. 
Based on the definition of mutual information on 3D Gaussian splatting, \cite{tao2024rt} approximates uncertainty with the magnitude of the parameter updates of the map, which is then used for estimating the information gain of candidate viewpoints.

\subsection{3D Gaussian Splatting}
3D Gaussian splatting proposed in \cite{kerbl20233dgs} provides a unique representation that captures geometry with Gaussians while encoding color and opacity information for high-quality rendering of scenes.
Building upon this representation, several approaches focus on real-time construction of such maps with color and depth measurements in simultaneous localization and mapping (SLAM) frameworks.
The method in \cite{keetha2024splatam} employs a silhouette mask on rendering for efficient optimization of the scene.
The work in \cite{matsuki2024gaussian} addresses the monocular SLAM problem with 3DGS along with analytical Jacobians for pose optimization.
The authors of \cite{peng2024rtg} propose handling Gaussian parameters differently for color and depth rendering by more explicitly representing surfaces for depth rendering, resulting in a more memory-efficient representation of the scene.
Similarly, \cite{hu2025cg} estimates stability and uncertainty of the Gaussians for efficient representation of the scene.
An advantage of the explicit representation of the Gaussian points is the ease of collision checking.
In comparison with other works, most notably \cite{chen2024splat}, we handle collision avoidance as chance constraints - in our case the size of the confidence ellipsoids depends on the number of Gaussians in the environment.
There are also numerous methods that address the problem of scene understanding with language-embedded 3DGS.
These approaches typically obtain 2D language features from Contrastive Language–Image Pre-training \cite{radford2021learning} (CLIP) and other foundation models, and embed these image features into 3D Gaussians.
More recent vision foundation models such as AM-RADIO~\cite{Ranzinger_2024_CVPR} distill complementary capabilities of multiple vision foundation models into a unified visual representation, providing an alternative feature backbone for open-vocabulary scene understanding.
Methods in \cite{shi2024language,qin2024langsplat,liao2024clip} use scene-specific autoencoders or quantization to obtain compact representations of the language features.
Other approaches \cite{yu2024language,zuo2024fmgs} leverage feature fields similar to the method proposed in \cite{kerr2023lerf} for Neural Radiance Fields (NeRFs).
In this work, we focus on an explicit representation that is amenable to planning.
We associate every Gaussian point in the map with a language embedding that is subsequently compressed using its principal components, yielding a scene-agnostic language-embedded map.

A sub-class of approaches \cite{zhu2024loopsplat, yugay2023gaussian} processes the scene as submaps by only optimizing the Gaussians on the submap level.
The approach in \cite{zhu2024loopsplat} performs submap alignment using multi-view pose refinement on keyframe images. 
Similar to these methods, we use a submapping approach for efficient storage of the map, focusing instead on the problem of optimizing Gaussians across multiple submaps to facilitate large-scale navigation.

\subsection{Semantic Mapping and Task-based Navigation}
Metric-semantic maps, which integrate both geometric and semantic information, provide actionable and informative representations of the environment.
Common forms of metric-semantic maps include semantics-augmented occupancy maps~\cite{asgharivaskasi2023semantic, dang2018autonomous}, object-based semantic maps~\cite{liu2024slideslam} and 3D scene graphs~\cite{armeni20193d, wu2021scenegraphfusion, looper20233d, hughes2024ijrr}. Among these, 3D scene graphs have emerged as a powerful representation, capable of capturing broader semantic concepts and the underlying contextual relationships within environments~\cite{armeni20193d}.
These models provide a compact, symbolic representation of semantic entities in the environments and their relationships~\cite{wu2021scenegraphfusion, looper20233d,hughes2022hydra}.
Other approaches incorporate additional information such as free space, object detections, and room categories on top of geometric representations~\cite{hughes2024ijrr}. 
This was extended by \cite{bavle2023s} to incorporate structural elements (e.g., walls) and clustering rooms.
Large Language Models (LLMs) have also been utilized to infer semantic relationships in the hierarchy~\cite{Strader24ral-autoAbstr} where structure is not readily available. 
To further generalize the use of scene graphs and capture a broader range of concepts for more complex robotic tasks, language features have been integrated to enable \emph{open-set} scene understanding. 
Clio~\cite{maggio2024clio} constructs a hierarchical scene graph, where the set of objects and regions is inferred from the list of given tasks.
Similarly, OrionNav~\cite{devarakonda2024orionnav} leverages open-set segmentation and LLM-based room clustering to build open-vocabulary scene graphs.
ConceptGraphs~\cite{conceptgraphs} constructs open-vocabulary 3D scene graphs from RGB-D image sequences by leveraging 2D foundation models for instance segmentation and projecting semantic features into 3D point clouds. 
It fuses multi-view information to create 3D objects annotated with vision and language descriptors, and uses large vision-language models to generate object captions and infer inter-object relations, resulting in comprehensive 3D scene graphs. 
HOV-SG (Hierarchical Open-Vocabulary 3D Scene Graph)~\cite{werby2024hierarchical} extends these capabilities to large-scale, multi-story environments. 
It introduces a hierarchical structure encompassing floors, rooms, and objects, each enriched with open-vocabulary features derived from pre-trained vision-language models.
CURB-OSG~\cite{steinke2025collaborative} constructs open-vocabulary dynamic 3D scene graphs for large-scale urban environments by fusing camera and LiDAR observations from multiple agents through a collaborative SLAM framework.
It constructs a hierarchy comprising roads and intersections, static objects, and dynamic road users.
ATLAS focuses on single-robot task-driven navigation, embedding dense language features directly within its geometric representation and constructing a sparse semantic hierarchy without relying on object-level segmentation priors.
In addition, while these methods that obtain an explicit semantic label at each level of a sparse, hierarchical representation capture the structure of known environments, such approaches cannot be directly applied to unknown and unstructured environments.
We instead try to find a balance between the sparsity provided by the hierarchical representation in scene graphs and the flexibility provided by dense, continuous feature embeddings.

\subsection{Task and motion planning}
A fundamental challenge in task-driven planning (beyond mapping and exploration) is the identification of objects and following paths likely to result in localizing the object in the minimum possible time.
~\cite{papatheodorou2023finding} proposed a planner that prioritizes the discovery of objects of interest and ensures their complete and high-resolution reconstruction. 
Learning based approaches that estimate semantic maps beyond line of sight have also been proposed ~\cite{georgakis2021learning} utilizing confidence in semantic labels from unobserved regions to guide the search.
Recent work has also focused on leveraging hierarchical scene graphs for semantic search tasks~\cite{dai2024optimalscenegraphplanning} using LLMs to guide an optimal hierarchical planner for semantic search, although these approaches assume that the map is available \textit{a priori}.
Ray et al.~\cite{ray2024task} formulate task and motion planning directly over such representations, using the hierarchy to construct sparse planning problems and incrementally introduce task-relevant objects to improve scalability.
In contrast, ATLAS incrementally constructs its metric-semantic map while exploring an initially unknown environment and couples hierarchical semantic planning with dense Gaussian geometry for collision-aware navigation.

Partially Observable Markov Decision Processes (POMDPs) have also been considered for planning on partially unknown scene graphs~\cite{amiri2022reasoning}. 
Reinforcement Learning based methods ~\cite{ravichandran2022hierarchical} have also been used to leverage scene graphs and graph neural networks to compute navigation policies, biasing the search toward regions of interest.
Vision Language Frontier Maps (VLFM)~\cite{yokoyama2024vlfm} utilized a Vision-Language Model (VLM) for estimating the similarity between the task and images projected onto a bird's eye view to score frontiers by their relevancy.
Large Language Models have also been directly used to select robot behaviors from a fixed library to select actions that minimize the time required to complete a task~\cite{ravichandran2024spine}.
While these methods focus entirely on completing the task with minimal maps along the way, we instead aim to generate a rich map that is reusable across a variety of tasks.

\section{Methods}
\label{sec:method}

\begin{framed}
ATLAS addresses the problem of exploring an unknown environment to complete a navigation task specified in natural language (e.g. ``find the parking lot'') while minimizing a cost that is a combination of the time required to complete the task and the control effort. It is designed to enable robots that operate in cluttered indoor scenes as well as large-scale outdoor environments while running in real time on SWaP-constrained platforms equipped with only vision sensors.
\end{framed}

This section develops the problem formulation given above while discussing the two key modules of our system.
\begin{enumerate}
\item \textbf{Mapping:} \cref{sec:mapping} discusses how ATLAS incrementally builds and maintains two kinds of representations of the environment for task-based navigation. The first one is a sparse, hierarchical and semantic representation that can be adapted for different tasks in real time. The second is a dense geometric representation that is used for collision-free planning.
\item \textbf{Planning:} \cref{sec:planning} discusses how this representation is used to build a two-level planner. A graph-based planner uses a task-based semantic relevance score recorded in the map to identify sequences of waypoints that the robot must visit. A continuous trajectory planner uses motion primitives to refine these high-level paths into dynamically feasible trajectories for a wheeled robot while ensuring that there are no collisions.
\end{enumerate}
The mapping and the planning modules interact in a closed-loop manner to enable the robot to execute semantically-informed actions to acquire new information that is relevant for completing the task.

We build our approach upon several existing mapping and planning methods in a modular manner. We use CLIP-DINOiser~\cite{wysoczanska2025clip} for computing language features, build upon SplaTAM~\cite{keetha2024splatam} for Gaussian splatting, and leverage prior work in motion primitive-based planning~\cite{tao2024rt,liu2017mpl} for low-level continuous trajectory planning in the Gaussian splatting map.
We adopt these methods in our implementation, though alternative approaches for each module can also be used.

\subsection{Hierarchical Semantic Perception}
\label{sec:mapping}

\begin{figure*}[!tpb]
    \centering
    \includegraphics[width=0.8\linewidth]{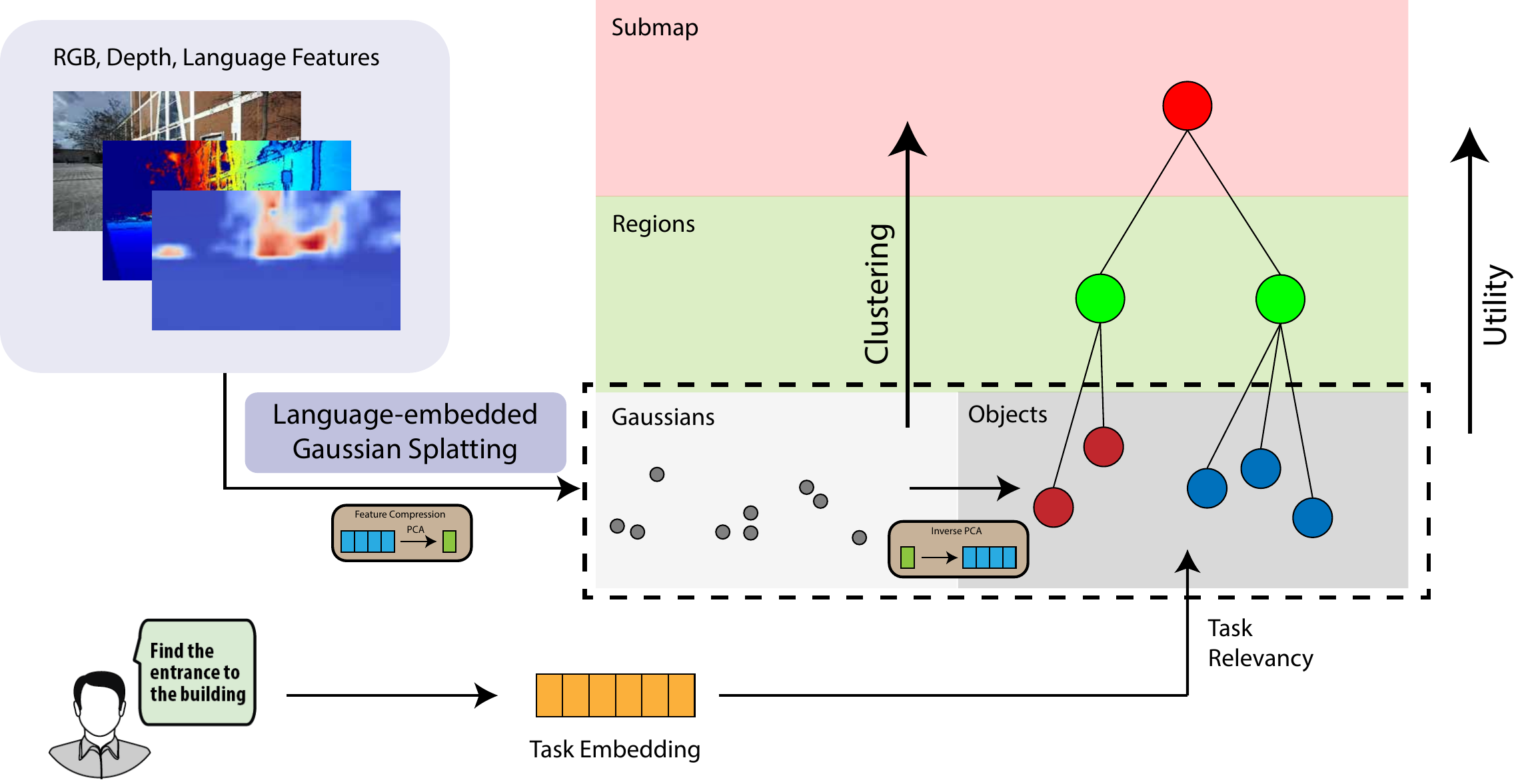}
    \caption{\textbf{Overview of the hierarchical semantic perception module.} RGB, depth and language features are embedded into an explicit representation that is fit using Gaussian splatting. Language features are compressed using principal component analysis (PCA). A hierarchical metric-semantic representation is constructed by agglomerative clustering of the Gaussians in the map. Given a task embedding, relevance scores are computed for the Gaussians (colored blue to red in increasing relevance) and propagated up the tree to result in region-level relevance scores.}
    \label{fig:perception_diagram}
\end{figure*}

The overall idea is to learn an explicit representation of the environment using Gaussian splatting along with language-based attributes for each location.
As we discussed in \cref{sec:related_work}, there are many existing approaches in the literature to build such representations. ATLAS makes two key innovations in this context.
First, we use agglomerative clustering to fuse dense low-level details of the environment and build a hierarchical representation---this is necessary for operating over large regions. And second, we develop a ``submapping'' module that can switch local maps (with semantic as well as metric information) in and out of our navigation stack to enable operation over very large areas, e.g., while the global map might have a few million Gaussians, the planner only needs to work with a few hundred thousand of them within each submap. \cref{fig:perception_diagram} gives an overview of the hierarchical semantic perception module.

\subsubsection{Language-embedded 3D Gaussian splatting}
\label{ssub:language_embedded_3d_gaussian_splatting}

We use RGB images and depth information to build an explicit representation of the environment as a set of ellipsoids (often called ``Gaussians'') each with a specific location, color and density that are optimized using Gaussian splatting (3DGS) \cite{kerbl20233dgs}. Suppose the map
\beq{
    \xi : x \mapsto (c, \s, o, f)
    \label{eq:map}
}
where the argument $x \in \reals^3$ is the location of each Gaussian, $\s \in \reals_+$ is an isotropic variance, $c \in \reals_+^3$ is the RGB color, and $o \in \reals_+$ corresponds to the opacity of each Gaussian. Gaussian splatting approaches store this map $\xi$ only at a few specific locations, say $\{x_1, x_2, \dots, x_N\}$, unlike approaches based on neural radiance fields (NeRFs) which store the function $\xi$ as a neural network. The fourth quantity $f \in \reals^m$ in \cref{eq:map} is an $m$-dimensional language-guided feature vector. As shorthand, we will use
\[
    c(x), \s(x), \dots
\]
to denote the color or variance of the Gaussian stored in the map $\xi$ at location $x$.

\medskip \textbf{Fitting a Gaussian splatting representation.}
The map $\xi$ can be used to render an RGB image from a new viewpoint $(x_0, t_0)$ where $x_0 \in \reals^3$ is the camera center and the direction $t_0 \in \reals^3$ corresponds to a pixel $u \in \integers^2$ on the image plane. Suppose we denote the locations of the Gaussians in our stored map $\xi$ that intersect the ray $t_0$ by $(x_1, \dots, x_n)$, and suppose that they are arranged in ascending order of their distance from the image plane. Then, using the volume rendering equation, the rendered color at a pixel $u$ is the average color of the Gaussians along this ray, each weighted by the probability that the ray travels to that point without encountering a solid object. More formally,
\[
    \aed{
    \reals^3 \ni c_\pi(u) &= \sum_{i=1}^n c(x_i) \a(x_i, u) \prod_{j=1}^{i-1} (1-\a(x_j, u)), \text{where}\\
    \a(x, u) &= o(x) \exp(-\f{1}{2 \s^2_\pi(x)} (u - x_\pi)^\top (u - x_\pi)).
    }
\]
The shorthand $\pi \equiv \pi(x_0, t_0)$ and $x_\pi$ denotes projection of the point $x$ upon the image plane for a viewpoint $(x_0, t_0)$, and similarly $\s_\pi(x)$ denotes the projection of a ball of radius $\s(x)$ upon this image plane. Features $f \in \reals^m$ can be projected upon the image plane using a similar expression as the one above, with $c(x_i)$ replaced by $f(x_i)$. The depth at pixel $u$ can also be calculated using a very similar expression, one averages the distance along the ray instead of the color $c(x)$.

A Gaussian splatting-based map $\xi$ is fit to the data by regressing these rendered colors against the observed color intensities $c^*(u)$ at each pixel, e.g., the L1 loss takes the form
\[
    \ell_c(\xi; \pi) = \sum_{u \in \Om} \norm{c_\pi(u) - c^*(u)}.
\]
We use a similar loss $\ell_f$ to fit the features $f$.
We discuss shortly how the ground truth features are computed. For incremental mapping applications such as ATLAS, given the latest RGB image and depth measurements (say, from a LiDAR or a depth/stereo camera), one initializes Gaussians in the 3D physical space by back-projecting pixels using their depth and color. The entire map $\xi$ is then refined using a few iterations of stochastic gradient descent. When depth measurements are available (they are in ATLAS), the estimated depths are also regressed against the observed depths of these Gaussians.

\begin{figure}[tb]
\centering
\includegraphics[width=\linewidth]{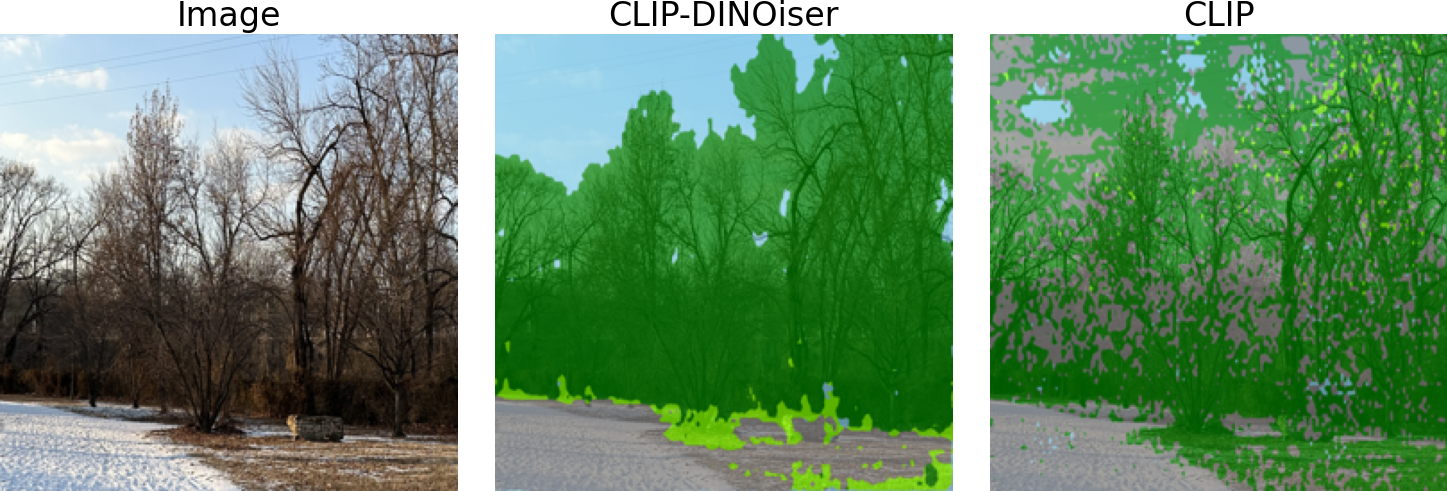}
\caption{\textbf{ATLAS uses CLIP-DINOiser to obtain fine-grained open-vocabulary semantic features from images.} CLIP-DINOiser refines the contextual information provided by CLIP features with the local structure captured in DINO features. This is particularly important for obtaining dense semantics of unstructured scenes. The above segmentation is obtained from the prompt [`trees', `grass', `snow', `sky'].}
\label{fig:clip_comparison}
\end{figure}

\medskip \textbf{Language-based features.}
Let us now discuss how the features $f(x)$ are optimized. ATLAS computes the ground-truth features $f^*(u)$ densely at each pixel $u \in \Om$ using a lightweight approach called CLIP-DINOiser~\cite{wysoczanska2025clip} which refines MaskCLIP~\cite{dong2023maskclip} features by incorporating localization priors extracted from DINOv2~\cite{oquab2023dinov2}.
These features are typically more fine-grained than, say, multi-scale features obtained from CLIP and other foundation models because DINO features are more sensitive to local structures in the image compared to CLIP.
A visualization of these features is provided in \cref{fig:clip_comparison}.
Alternative feature backbones could also be integrated within this framework.
We use CLIP-DINOiser in our implementation due to its lightweight dense feature extraction and fine-grained localization.

\begin{figure}[tpb]
\centering
\begin{tikzpicture}[scale=0.65]

\begin{axis}[
    xlabel={PCA Dimension},
    ylabel={Cosine Similarity},
    ytick={0.92, 0.94, 0.96, 0.98, 1},
    ymin=0.90,
    ymax=1.01,
    ybar,
    bar width=15pt,
    symbolic x coords={12,24,48,128,256,384},
    xtick=data,
    ymajorgrids=true,
    grid style=dashed,
    axis y line*=left,
    axis x line*=bottom,
    label style={font=\normalsize},
    width=13cm,
    height=6cm,
]

\addplot coordinates {
    (12, 0.9300597067995255)
    (24, 0.9529455201139466)
    (48, 0.9708794718075209)
    (128, 0.9889871331096114)
    (256, 0.9966775355476528)
    (384, 0.9992471339162317)
};

\end{axis}
\end{tikzpicture}
\vspace{-0.8cm}
\caption{\textbf{Cosine similarity between original and re-projected features on ScanNet.} PCA is an approximate isometry, i.e., it preserves pairwise Euclidean distances between points from the original space in the embedding space, and does so exactly when the dimensionality of PCA features matches that of the original data. CLIP features are normalized to have unit $\ell_2$-norm. As we see, the cosine similarity of the embeddings of CLIP features (after projecting them back to the original space), against the original CLIP features is rather high even when the dimensionality is reduced from $m=512$ to $m=24$.}
\label{fig:pca}
\end{figure}

Language-based features $f(x) \in \reals^m$ are typically high-dimensional vectors, e.g., $m = 512$ in CLIP. Storing such features densely in the scene for every Gaussian requires a lot of space, e.g., in a large scene with 10M Gaussians the space required to store features would be more than 20 GB. Even if one were to store them, real-time computations with the features are difficult.
For example, for an HD camera with resolution 1920 $\times$ 1280, even with just 10 Gaussians along each ray, more than 24M $m$-dimensional feature vectors need to be projected upon the image plane for each rendering, which increases the computational complexity of backpropagation while fitting Gaussian splatting.
It is necessary to compress language-based features to a lower-dimensional space before storing them in the map. For example, many existing approaches use features fields computed from multi-scale images \cite{kerr2023lerf, yu2024language} or scene-specific autoencoders \cite{qin2024langsplat}.
While the former leads to blurrier semantic features, the latter is specific to a given scene and therefore cannot be used for exploration/mapping.
ATLAS uses a simple strategy to work around this problem with Principal Component Analysis (PCA) to reduce the dimensionality of the CLIP-DINOiser features.
This ``encoder'' is fit offline by running PCA~\cite{ross2008incremental} on the COCO 2017~\cite{lin2014microsoft} dataset (we use incremental PCA to reduce memory usage). At runtime, CLIP-DINOiser features are projected into this lower-dimensional space using the PCA encoder and treated as the ground-truth features $f^*(u)$. For evaluating the relevance of a part of the scene to a given task at hand, these compressed features stored in the Gaussians can be decoded back into the higher-dimensional space easily. A nonlinear encoder-decoder pair, e.g., a variational auto-encoder, would lead to more substantial compression than PCA. But ATLAS uses this simple strategy because (i) it works well even in our large-scale experiments, and (ii) even if offline training of a nonlinear encoder-decoder pair is easy, inference on such a large number of vectors at runtime is difficult to implement on the limited computational budget available on a robot. \cref{fig:pca} shows a simple ablation experiment for CLIP-DINOiser features using PCA on the ScanNet dataset.

\subsubsection{Submapping}
\label{ssub:sub_mapping}

\begin{figure}[!tpb]
    \centering
    \includegraphics[width=\linewidth]{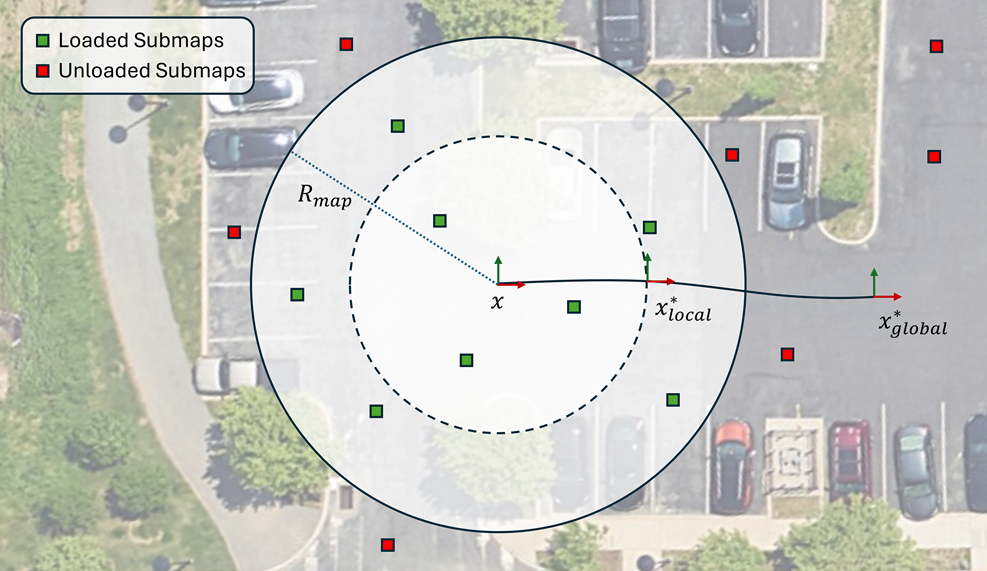}
    \caption{\textbf{An illustration of the parameters that are relevant to the submapping and collision checking}. The location of the robot is $x$, the local goal location is $x^*_{\text{local}}$ (as explained in \cref{sec:trajopt}) while  the global goal location is $x^*_{\text{global}}$. Submaps are loaded within a ball of radius $R_{\text{map}}$ around the robot.
    }
    \label{fig:submap}
\end{figure}

As we discussed, an explicit representation like 3DGS requires a large amount of memory when the size of the scene is large.
In ATLAS, we are interested in maps as large as $\sim$20,000 \unit{\square\m} with $\sim$10 M Gaussians. Maintaining this representation is prohibitive on SWaP-constrained platforms.
Therefore, we need a ``submapping'' procedure which can dynamically create, load, and unload small parts of the larger map as the robot explores the environment.
The map in ATLAS
\[
    \xi = {\textstyle \bigcup_i} \xi_i
\]
where $\xi_i: x \mapsto (c, \s, o, f)$ are individual submaps that store all the parameters associated with the Gaussians.
Each submap has an anchor pose (location and orientation in $\text{SE}(3)$)
\[
    \xi_i \equiv \xi_i(x^{\text{anchor}}_i)
\]
and locations of Gaussians are stored relative to this anchor pose (which is computed from onboard odometry). 
At each iteration, the robot only works with submaps that are within a predefined radius from its location in global coordinates, and offloads all other submaps off the GPU.
If there is no anchor pose within a predefined distance from the robot, a new submap is spawned.
We pick this distance heuristically to be 2--5 \unit{m} for indoor or outdoor experiments, with the idea that stereo vision-based depth measurements from the robot are accurate within this radius.  \cref{fig:submap} shows an illustration of this submapping procedure.

\begin{figure*}[!htpb]
    \centering
    \includegraphics[width=\linewidth]{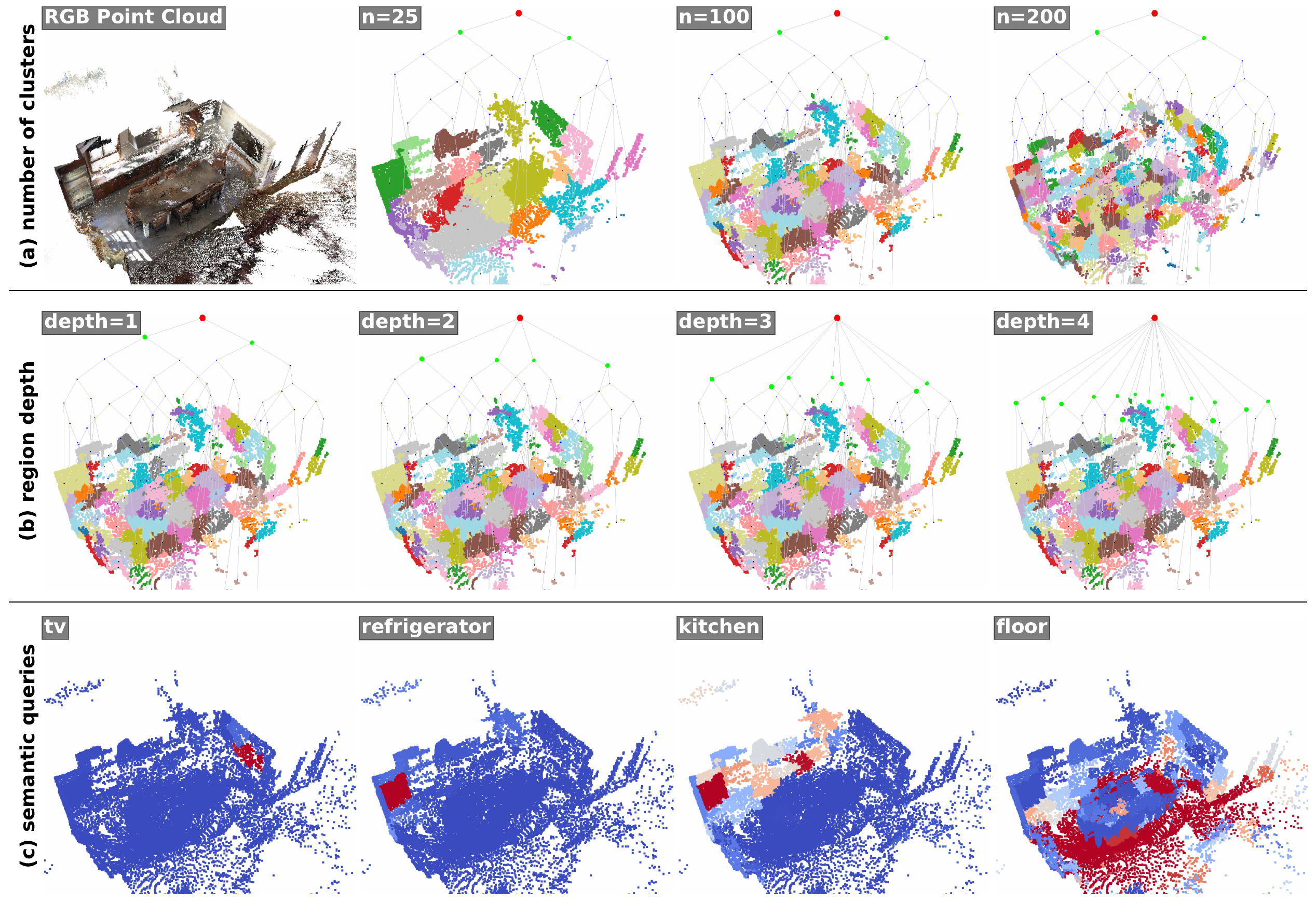}
    \caption{\textbf{Visualization of the agglomerative clustering on an indoor scene.} While our method is designed for unstructured outdoor environments, it also effectively captures detailed semantics in structured indoor settings. (a) Agglomerative clustering is applied to the dense language-embedded Gaussians in the map to identify object-level clusters. The individual clusters are each denoted by a different color. Increasing the number of clusters produces a more detailed segmentation of the scene but also yields a larger tree. (b) The agglomerative structure allows us to build a hierarchical representation of the submap at any desired resolution, trading off spatial and semantic resolution with compute efficiency for querying submaps. (c) The object clusters retain semantic information which can be queried directly. The clusters are constructed from a dense representation which provides the flexibility of representing objects such as `tv' and `refrigerator' while also being able to accommodate non-object elements like `floor' and more abstract concepts like `kitchen'. The relevancy of each cluster to the task query is colored blue (low) to red (high). }
    \label{fig:agglomerative}
\end{figure*}

In contrast to traditional voxel-based methods that work with a fixed discretization, our design allows flexibility in terms of storage because it does not pre-allocate memory for the Gaussians. This flexibility can be used to build denser maps in cluttered environments and sparser maps in large open spaces. Another useful aspect of the submap-based design is that each individual submap is tightly coupled to its anchor point, while the anchor points are loosely coupled to each other. If a localization and mapping (SLAM) algorithm requires a particular anchor pose to be corrected (e.g., after loop-closure is detected), ATLAS need only update the anchor pose but it does not need to update the Gaussians. This ensures that the global map is quite consistent across large-scale operation even if loop-closures are not propagated to the entire map.
In our implementation, we benefit from such loop-closures detected in the onboard ZED 2i odometry system.

\subsubsection{Metric-semantic clustering}
\label{ssub:metric_semantic_clustering}

Semantic information in the environment is useful in large-scale environments, especially in situations where the task is specified in natural language, because language-guided features computed on observed images can be correlated easily with such specifications. A popular way of achieving this is to use image segmentation and back-project these segmentations upon the map~\cite{conceptgraphs, maggio2024clio, werby2024hierarchical}. This is inadequate for two reasons. First, it is difficult to implement robustly in practice.
Small inaccuracies in masks or occluded objects can result in large inaccuracies in characterizing the semantic information in the 3D map. Second, the resultant 3D representation does not characterize (i) how objects are semantically related to other objects in their neighborhoods, or (ii) the semantic features of large areas. This makes it difficult to use the representation for large-scale tasks.

\medskip \textbf{Agglomerative metric-semantic clustering}
To work around these issues, in ATLAS we build a metric-semantic representation using an agglomerative clustering of the Gaussians using a distance function that depends on both their locations and the similarity of their language-guided features. Formally, the distance is
\[
    d(x,x') = \|x-x'\|_2 + \lambda \left(1 - \frac{f(x)^\top f(x')}{\|f(x)\|\,\|f(x')\|}\right)
\]
for a hyper-parameter $\l \geq 0$. Note that $f(x)$ are the language-guided features after compression via PCA.
Clustering is performed within a submap when it is offloaded from the GPU (as we discussed in the previous section).
The root of the tree corresponds to the submap.
Objects correspond to the leaf nodes of the clustering.
The agglomerative data structure allows us to define ``regions'' at any depth in the tree depending on the desired resolution.

\cref{fig:agglomerative} shows how increasing the number of clusters results in a more detailed segmentation of the scene. The agglomerative clustering is able to segment out individual objects such as chairs and the refrigerator without the need for image segmentation priors. Different semantic regions can be defined at different depths on the same tree, with features that are consistent with those of the leaves. In experiments, we observed that such agglomerative semantic features do retain accurate semantic information of the scene.

For a real-time implementation, we sample $\sim$5,000 Gaussians from the submap and use a tree of depth three. The number of leaves (object clusters) is limited to 100 and the degree of the root node is limited to 4 for outdoor scenes. This strategy, along with the advantages of a dense representation, provides a flexible way to represent both individual objects and broader semantics across larger regions.

\medskip \textbf{Computing task similarity in the map using the agglomerative tree}
Given a task specified in natural language as ``Navigate to the entrance of the pier'', we would like to compute the similarity in embedding-space of the text features against the features stored in the agglomerative tree.
To do so, we first retrieve dense features of each object node in the tree (leaves) using inverse PCA and compute the cosine similarity.
We propagate object-level relevancy up the tree to obtain similarity scores for all region- and submap-level nodes in the tree.
The similarity score of each node is the sum of the scores of its children.
Finally, the score of the root represents the score of the entire submap for the task. This submap-level task relevancy information is also stored in each submap in addition to the Gaussian splatting parameters.
Within the local map of loaded submaps, rather than reconstructing the agglomerative tree each time a submap is updated, we instead bin the Gaussians in the local map in a 0.1 \unit{\m} spatial grid and compute relevancy on the aggregated features.

\subsection{Hierarchical Planning}
\label{sec:planning}

The planning stack in ATLAS operates at two levels. The first level is a discrete planner that can do long-term semantic reasoning using the agglomerative clustering within each submap and across multiple submaps. This level identifies a sequence of local goal locations that may be followed to satisfy the task. The second level is a continuous trajectory planner that works in $\text{SE}(2)$ to compute collision-free trajectories to the current local goal location.

\subsubsection{Discrete planner}
\label{ssub:discrete_planner}

Given the hierarchical semantic map of submaps and regions, we aim to determine an optimal sequence of vantage points from which a robot can obtain observations to gather task-relevant information. Each vantage point corresponds to a node in the agglomerative tree with its associated score. We associate a travel weight with each pair of vantage points that represents the cost of moving between them, in this case defined by the Euclidean distance between the nodes. Our goal is to find a path that maximizes the cumulative score while ensuring that the total travel cost remains within a predefined budget.

In principle, this problem can be solved as a resource-constrained path selection problem over a graph whose nodes are the union of the nodes of the agglomerative trees of each submap.
In our largest experiments there are 379 submaps, each with four nodes.
If we associate travel weights across even a small fraction of these nodes, say the ones within some fixed Euclidean distance from each other, the resultant graph can be extremely large for real-time operation on a SWaP-constrained robot.
To address this, we adopt a hierarchical approach. At the coarse level, we construct a graph whose nodes represent submaps.
Two submap nodes are connected by an edge if the distance between them is below a specified threshold.
As discussed in \cref{ssub:metric_semantic_clustering} the root node in each submap also maintains a score of its relevance to the task.
On this coarse graph, we compute a path that maximizes the total submap utility weighted by distance, while staying within a fixed path‑length budget, producing an ordered sequence of submaps that respects the travel limit and prioritizes highly relevant regions.
The search is implemented using a priority queue over partial paths, ordered by their accumulated utility and pruned whenever the travel budget would be exceeded.

Given this submap sequence, we then refine the plan within and between the selected submaps. We construct a denser graph over the region nodes of the chosen submaps and connect nodes according to the high‑level order.
On this refined graph, we run Dijkstra’s algorithm to compute optimal paths between and within submaps, producing a reference path that follows the prescribed high‑level plan.

\subsubsection{Continuous trajectory planning using motion primitives}
\label{sec:trajopt}

We use ground robots in our experiments and use a unicycle dynamics model for trajectory planning. The state of the robot $x \in \text{SE}(2) \cong \reals^2 \times S^1$. The height above the ground plane is set to zero. The control inputs $(v, \omega)$ are the translational and rotational speed, respectively. Formally, the dynamical system is
\[
    \dv{x}{t} = \bmat{v \cos (x_3)\\ v \sin (x_3)\\ \omega}.
\]
Suppose the discrete planner prescribes a local goal region $x^*_\text{local} \subset \text{SE}(2)$. The continuous trajectory planning problem seeks to identify a dynamically feasible trajectory that obeys pre-specified constraints on the maximal translational and angular velocity while minimizing a cost that encodes time to travel and the control effort.

Even if the continuous trajectory problem is only in three dimensions, obstacles corresponding to the large number of Gaussians in the map make it challenging. We adopt a search-based procedure motivated by the work of Liu et al.~\cite{liu2017mpl}. We compute motion primitives corresponding to fixed control inputs for a time $\Delta t$. Primitives are computed for a grid of translational and rotational velocities that spans their maximal values. The cost of each primitive computed using $(v, \w)$ is
\[
   \l \Delta t + (v^2 + \omega^2) \Delta t,
\]
for a hyper-parameter $\l$. This leads to a graph whose nodes are terminal states for these primitives. Each edge on the graph of motion-primitives should be checked for collisions against the obstacles in the environment, e.g., by discretizing the edge at some fixed interval. We will describe an efficient procedure to do so subsequently. Any sequence of nodes on this graph is then a dynamically-feasible trajectory for the robot. We use the A* algorithm to find the best sequence. This continuous trajectory planner runs in a receding horizon fashion to update the trajectory with a fixed frequency, the local goal region $x^*_\text{local}$ may also be updated periodically.

\begin{figure}
    \centering
    \includegraphics[width=0.9\linewidth]{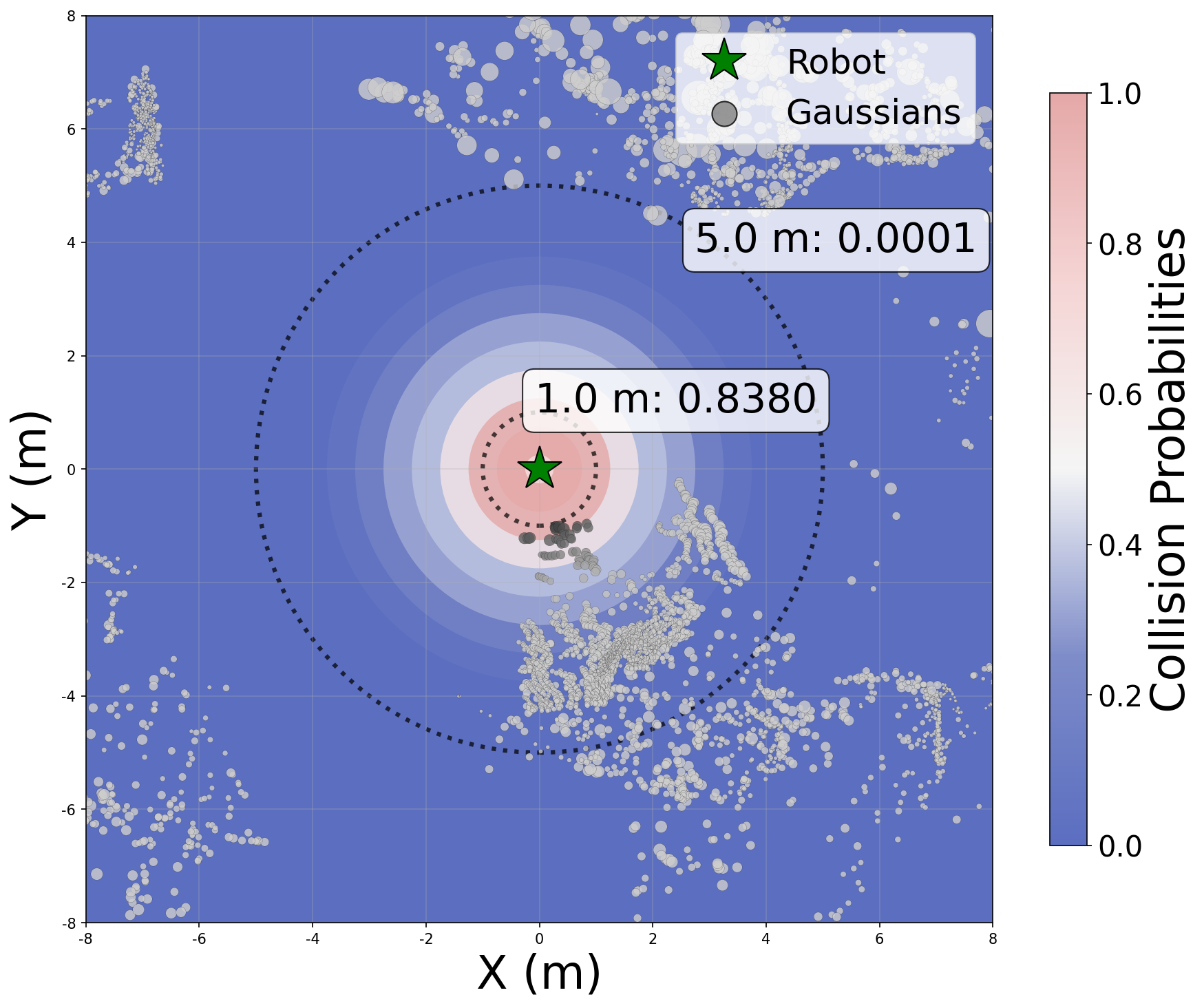}
    \caption{
    \textbf{A heatmap of collision probabilities as a function of distance from the robot (located at the green star).} The dotted circle denotes a ball of radius $r^*$ corresponding to the radius of the robot $r_0 = 0.7$, inflation factor $c = 1$, density $\r = 1250\ \unit{\per\cubic\m}$ and collision probability $\delta = 10^{-4}$ for the larger dotted circle while $\delta = 0.838$ for the smaller dotted circle. White dots denote the Gaussians and their size denotes the standard deviation (size is scaled up for visualization purposes). As one goes away from the robot location, the collision probability decreases.
    }
    \label{fig:collision}
\end{figure}

\begin{table*}[!ht]
\centering
\footnotesize
\setlength{\tabcolsep}{5pt}
\begin{tabular}{p{3.0cm}p{4.0cm}p{2.7cm}p{5.6cm}}
\toprule
\textbf{Module} & \textbf{Hyperparameter} & \textbf{Value} & \textbf{Purpose} \\
\midrule
\multirow{3}{*}{Language features} & CLIP feature dimension & 512 & Feature dimension before compression \\
& PCA feature dimension & 24 & Compressed features stored in each Gaussian \\
& CLIP encoder & ViT-B/16 & Computes image and task embeddings \\
\midrule
\multirow{3}{*}{Mapping} & RGB/depth image resolution & 320 $\times$ 240 & Input resolution for mapping \\
& Maximum depth range & 5 \unit{m} & Maximum depth measurements \\
& Gaussian update rate & 1 \unit{\Hz} & Frequency for online map optimization \\
\midrule
\multirow{3}{*}{Submapping} & Indoor submap radius & 2 \unit{m} & Local submap size for indoor experiments \\
& Outdoor submap radius & 5 \unit{m} & Local submap size for outdoor experiments \\
& Local map radius $R_{\text{map}}$ & 10 \unit{m} & Loads nearby submaps for collision checking \\
\midrule
\multirow{8}{*}{Planning} & Robot radius & 0.7 \unit{m} & Robot footprint used for collision checking \\
& Collision probability tolerance $\delta$ & 0.001 & Chance-constraint threshold for collision checking \\
& Collision-check distance $r^*$ & 5 \unit{m} & Distance beyond which collision probability is bounded by $\delta$ \\
& Gaussian density $\rho$ & 1250 \unit{\per\cubic\m} & Density used to compute $r^*$ offline \\
& Inflation factor $c$ & $1.5$ & Inflates collision bounds \\
& Local goal distance & 5 \unit{m} & Goal distance threshold \\
& Discrete planner rate & 0.2 \unit{\Hz} & Frequency for global planner \\
& Continuous planner rate & 1 \unit{\Hz} & Frequency for continuous trajectory generation \\
\bottomrule
\end{tabular}
\caption{Hyperparameters and model choices used in ATLAS. Values are shared across experiments unless an indoor/outdoor distinction is specified.}
\label{table:hyperparameters}
\end{table*}

\medskip \textbf{Efficiently computing the probability of collision of the robot against the 3DGS representation.}
This submap-based design already ensures efficient operation here while significantly reducing the number of Gaussians that need to be considered. We can further improve upon it as follows. We next present a short calculation to obtain a numerical formula for the smallest radius $r^*$ around the robot that one needs to consider to ensure that the probability of collision is less than some pre-specified threshold $\delta$. Suppose the location of the robot is $x$ within a submap with Gaussians at locations $\{x_1, \dots, x_n\}$. We wish to compute
\beq{
    \aed{
    &\P\rbr{\exists i \leq n: \norm{x_i - x} \geq r^*, \text{collision}(x_i, x)}\\
    &\leq \sum_{i=1}^n \ind{\norm{x_i - x} \geq r^*} \P(\text{collision}(x_i, x)),
    }
    \label{eq:collision_prob}
}
where
\[
    \text{collision}(x_i, x) = \ind{\norm{Z} \leq c r_0}
\]
for a normal-distributed random variable $Z$ with mean $(x_i - x)$ and variance $(\s(x_i)^2 + r_0^2)\ I_{3 \times 3}$. The radius of the robot is $r_0$ and the standard deviation of the Gaussian in question is $\s(x_i)$. The inequality follows from a union bound. Let $\s^2 = \s_{\text{avg}}^2 + r_0^2$ where $\s_{\text{avg}} = n^{-1} \sum_i \s(x_i)$ is the average standard deviation of the $n$ Gaussians. Because $Z$ is a Gaussian random variable, we can rescale it to obtain a standardized random variable $Z_{\text{std}}$ that has zero mean and unit variance to see that
\beq{
    \aed{
    &\P(\text{collision}(x_i,x))\\
    &= \P \rbr{\abs{Z_{\text{std}} - \f{\norm{x_i-x}}{\s}} \leq \f{c r_0}{\s}},\\
    &\leq \P \rbr{\abs{Z_{\text{std}} - \f{r^*}{\s}} \leq \f{c r_0}{\s}},
    }
    \label{eq:zstd}
}
because $\norm{x_i-x} \geq r^*$ and the collision probability is non-increasing in distance. This probability can be calculated analytically for a given $r^*$. We set the inflation factor $c \approx 1.5$ in our experiments to ensure that the collision probability computed using this expression is not excessively large. If we assume that the density of the Gaussians in the map is $\r$, then the cardinality
\[
\abs{\{i \leq n: \norm{x_i - x} \geq r^*\}} = n - \frac{4}{3} \pi {r^*}^3 \r.
\]
As a result, we can compute the smallest $r^*$ numerically using bisection search to ensure
\beq{
    \aed{
    &\P\rbr{\exists i \leq n: \norm{x_i - x} \geq r^*, \text{collision}(x_i, x)}\\
    &\leq \rbr{n - \frac{4}{3} \pi {r^*}^3 \r} \P \rbr{\abs{Z_{\text{std}} - \f{r^*}{\s}} \leq \f{c r_0}{\s}} \leq \delta.
    }
    \label{eq:bound}
}
In practice, we set $\delta = 0.001$. This calculation can be conducted offline for different values of the densities $\r$. It can also be done when the Gaussians are non-isotropic, either by choosing an upper bound on the variance along each axis, or by doing a more refined calculation in place of \cref{eq:zstd}. As a consequence of this calculation, in \cref{fig:collision}, to check for collisions of the robot, we only need to consider $\sim$370,000 Gaussians while the full map has $\sim$2.5 M Gaussians.

\subsection{Implementation Details}
We provide platform and module implementation details, with a summary of parameter values provided in \cref{table:hyperparameters}.

\medskip \textbf{Robot platform.}
All our real-world experiments use a Clearpath Jackal robot platform. It is modified to carry an onboard computer with AMD Ryzen 5 3600 CPU, Nvidia RTX 4000 Ada SFF GPU and 32 GB of RAM. Additionally, the robot is equipped with a ZED 2i stereo camera for obtaining RGB, depth and pose measurements. We use the onboard visual-inertial odometry from ZED 2i in our experiments. The platform also has an Ouster OS1-64 LiDAR that is used to obtain a more accurate (``ground truth'') estimate of odometry for evaluation purposes.

\medskip \textbf{Mapping.}
We build upon the Gaussian splatting-based mapping approach presented by Keetha et al. in \cite{keetha2024splatam}, with three key modifications (i) language-guided features as described in \cref{ssub:language_embedded_3d_gaussian_splatting}, (ii) submapping procedure described in \cref{ssub:sub_mapping}, and (iii) pose estimates obtained from visual-inertial odometry. We compress the 512-dimensional CLIP features down to 24 dimensions using PCA for storing it in the map.
Submaps have a radius of 2 \unit{m} in indoor experiments and 5 \unit{m} in outdoor experiments.
All RGB and depth images are resized to 320 $\times$ 240 before using them, and depth maps are culled to a maximum range of 5 \unit{m}.
The Gaussians in the map are updated at 1 \unit{\Hz}.

\medskip \textbf{Planning}
Both discrete and continuous planning modules are implemented in Python and build upon the motion planning library of Liu et al.~\cite{liu2017mpl}. The radius of the robot platform is set to 0.7 \unit{m}, the collision probability $\delta = 0.001$ and $r^* = 5$ \unit{m} which was calculated using $\r = 1250$ \unit{\per\cubic\m} Gaussians in the map as shown in \cref{fig:collision}. The local goal is always at a distance of at most 5 \unit{m} from the robot, and we load submaps within a ball of radius $R_{\text{map}} = 10$ \unit{m} centered around the robot on the GPU (see \cref{fig:submap}). We adopted the idea of parallelized collision checks presented in Tao et al.~\cite{tao2024rt} to enable efficient continuous planning online. The discrete planner runs onboard the robot every 5 mapping iterations, i.e., at 0.2 \unit{\Hz}, while the continuous planner runs at 1 \unit{\Hz}.

\begin{figure}[!h]
    \centering
    \includegraphics[width=0.9\linewidth]{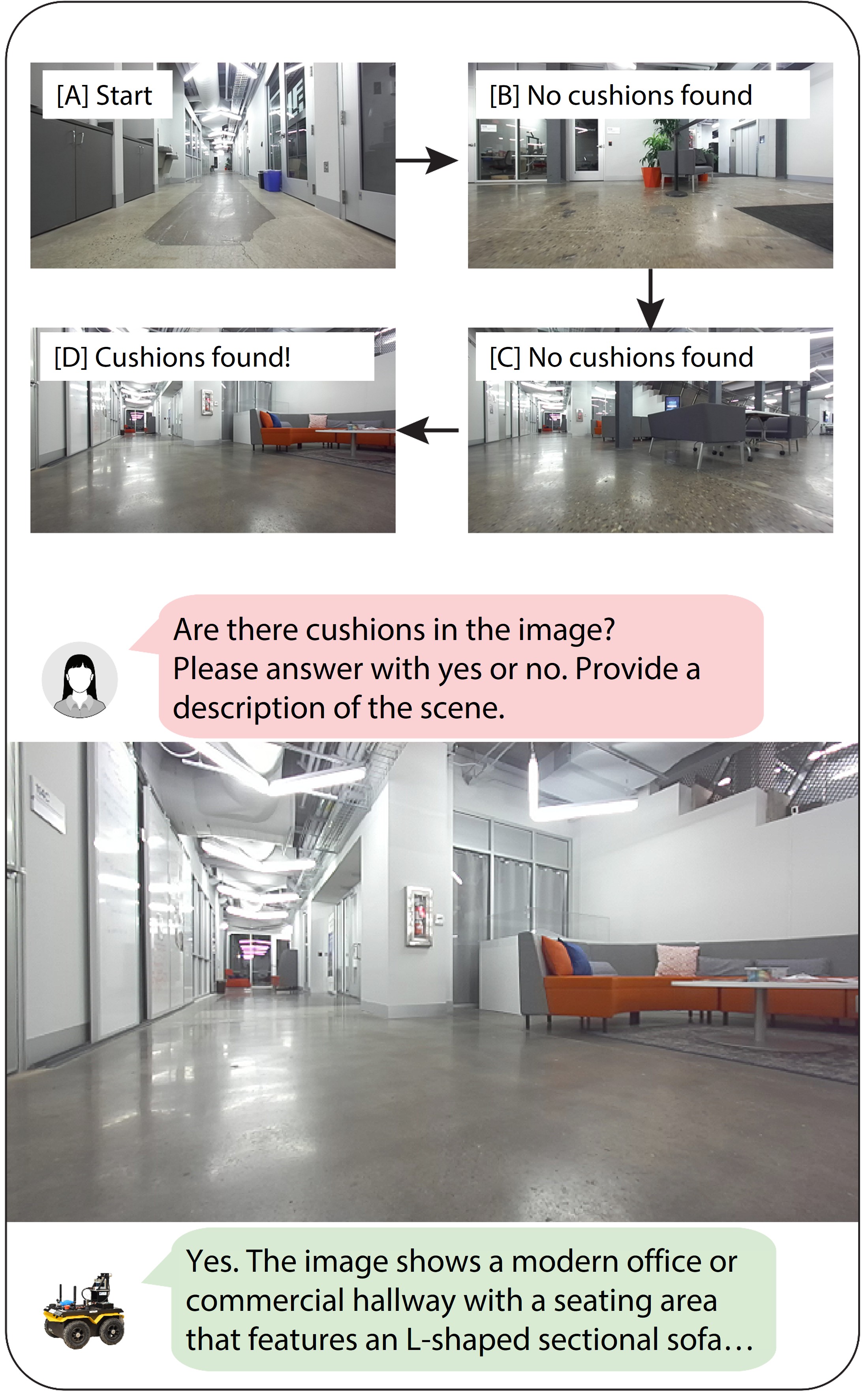}
    \caption{\textbf{Robot executing the task ``find cushions'' starting at [A].} The robot incrementally constructs a map with language-guided Gaussian splatting and identifies and navigates to regions in the map with high relevance. The VLM is queried at vantage points [B], [C], [D]. The images acquired at [B] and [C] have the second-order association of couches but no cushions. The task only terminates at [D] when cushions are found.
    }
    \label{fig:cushions_viz}
\end{figure}

\begin{figure}[!h]
    \centering
    \includegraphics[width=\linewidth]{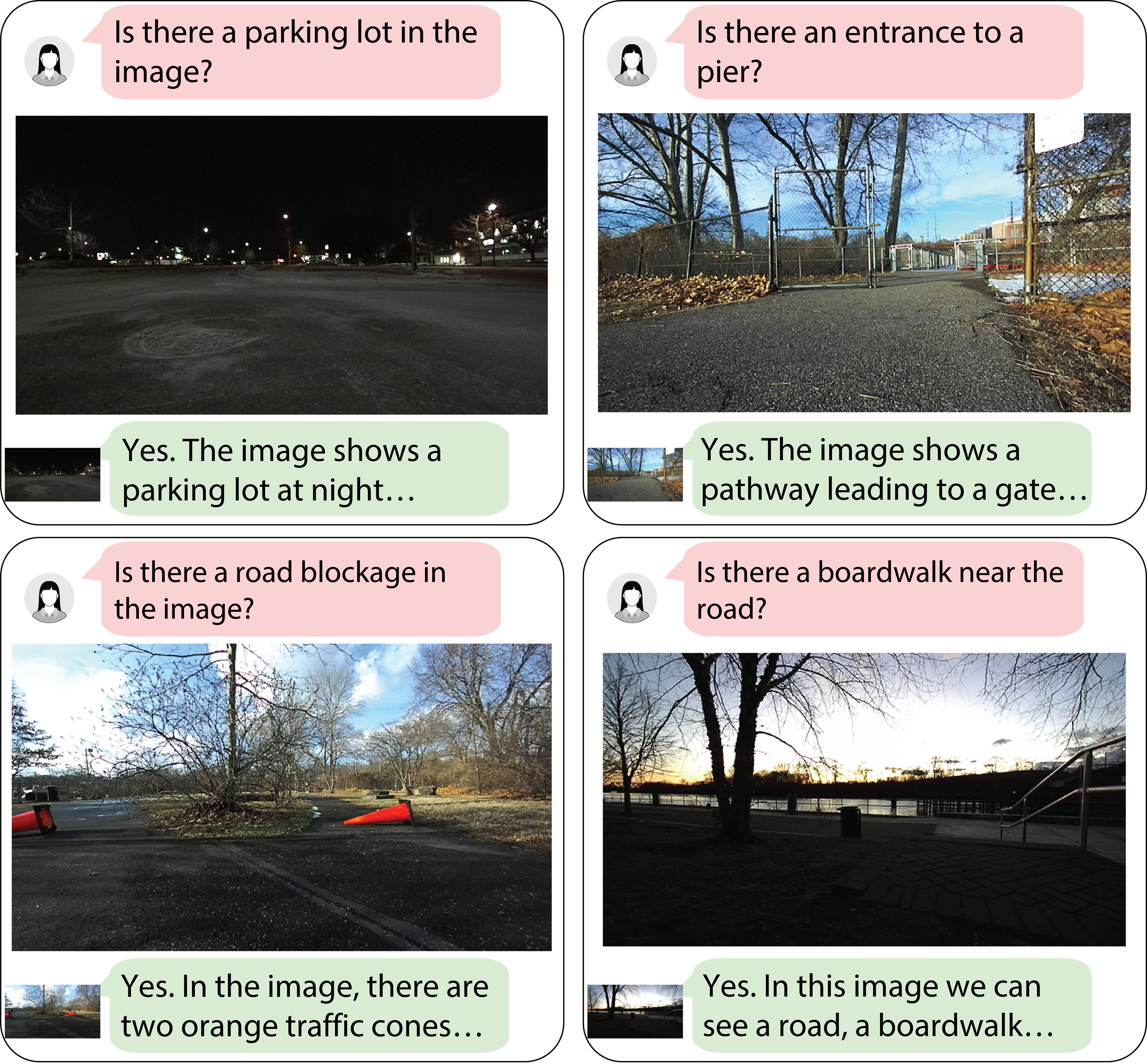}
    \caption{
    \textbf{Qualitative results showing the output of the VLM upon task termination.} The output for ``find boardwalk near road" task (top left), for ``inspect road blockage" task (top right), and for the ``find parking lot" task (bottom left), and for the ``navigate to entrance to pier" task (bottom right).
    }
    \label{fig:vlm_outputs}
\end{figure}

\medskip \textbf{Task specification}
The task is specified in natural language by the user, e.g., as a prompt ``Navigate to the entrance of the pier''. We use a ViT-B/16 encoder for the CLIP backbone. Text embeddings are computed from the task prompt using the CLIP text encoder~\cite{radford2021learning}. These text embeddings are used to compute the relevancy of different nodes in the agglomerative tree within each submap, as well as the relevance of different submaps using the features stored in their corresponding root nodes. Task re-specification is done by recomputing the task relevance scores and propagating them up the tree. This is one of the key features of ATLAS, the same submap can be efficiently re-queried for its semantic information as it pertains to different tasks.

\medskip \textbf{Task termination criteria}
We use LLaVA OneVision~\cite{li2024llava} to determine if the task has been completed. Given the language features of the current observation, we mask out pixels that are beyond the maximum depth range and compute the relevance of the remaining pixels against the task. If the relevancy is beyond a threshold, we query the VLM module with the image and the task. We ask it to reply with a yes or no and a description of the scene. One such example is visualized in \cref{fig:cushions_viz}.

\section{Experimental Evaluation}
\label{sec:robot-exp}

We discuss real-world robot experiments in different kinds of indoor and outdoor environments. These experiments are designed to test closed-loop autonomous navigation on a pre-built map and navigation without any prior map where the map is built incrementally over time.

\begin{table*}
\centering
\begin{adjustbox}{width=\linewidth}
\renewcommand{\arraystretch}{1.1}
\begin{tabular}{ccc rrrrr rr}
\toprule
\multirow{2}{*}{Experiment} & \multirow{2}{*}{Task} & Prior & \multicolumn{5}{c}{Distance (m)} & {Area} & {Num. of} \\
&  & Map & {PL (m)} & \multicolumn{2}{c}{SP (m / ratio)} & \multicolumn{2}{c}{GT (m / ratio)} & {(m$^2$)} & {Gaussians} \\
\midrule
\textbf{Indoor}\\
1 & Find cushions & \ding{56} & 72.53 & 34.87 & 0.48 & 29.72 & 0.41 & 973.28  & 0.63M \\ 
2 & Find plants, Exit building & \ding{56} & 44.55 & 31.60 & 0.71 & 30.16 & 0.68 & 973.28 & 0.45M \\ 

\midrule
\textbf{Outdoor}\\
1 & Navigate to entrance of the pier & \ding{52} & 185.78 & -- & -- & 184.86 & 0.99 & 3346.50 & 2.66M \\ 
2 & Inspect road blockage & \ding{56} & 94.25 & 45.16 & 0.48 & 43.52 & 0.46 & 1063.57 & 1.09M \\ 
3 & Find parking lot & \ding{56} & 69.19 & 46.439 & 0.67 & 43.22 & 0.62 & 1280.10 & 1.47M \\ 
4 & Find river near road & \ding{56} & 742.42 & 473.64 & 0.64 & 472.55 & 0.64 & 17791.14 & 7.77M \\ 
5 & Find boardwalk near road & \ding{56} & 1257.53 & 688.90 &  0.55 & 671.76 &  0.53 & 21870.34 & 11.28M \\ 
\bottomrule
\end{tabular}
\end{adjustbox}
\caption{\textbf{Overview of robot experiments.} PL path length, SP refers to the shortest path computed on the full planning graph after the experiment is complete, and GT refers to the ground truth computed manually using GPS information and Google Earth.}
\label{tab:real_expts}
\end{table*}

\begin{figure}
    \centering
    \includegraphics[width=\linewidth]{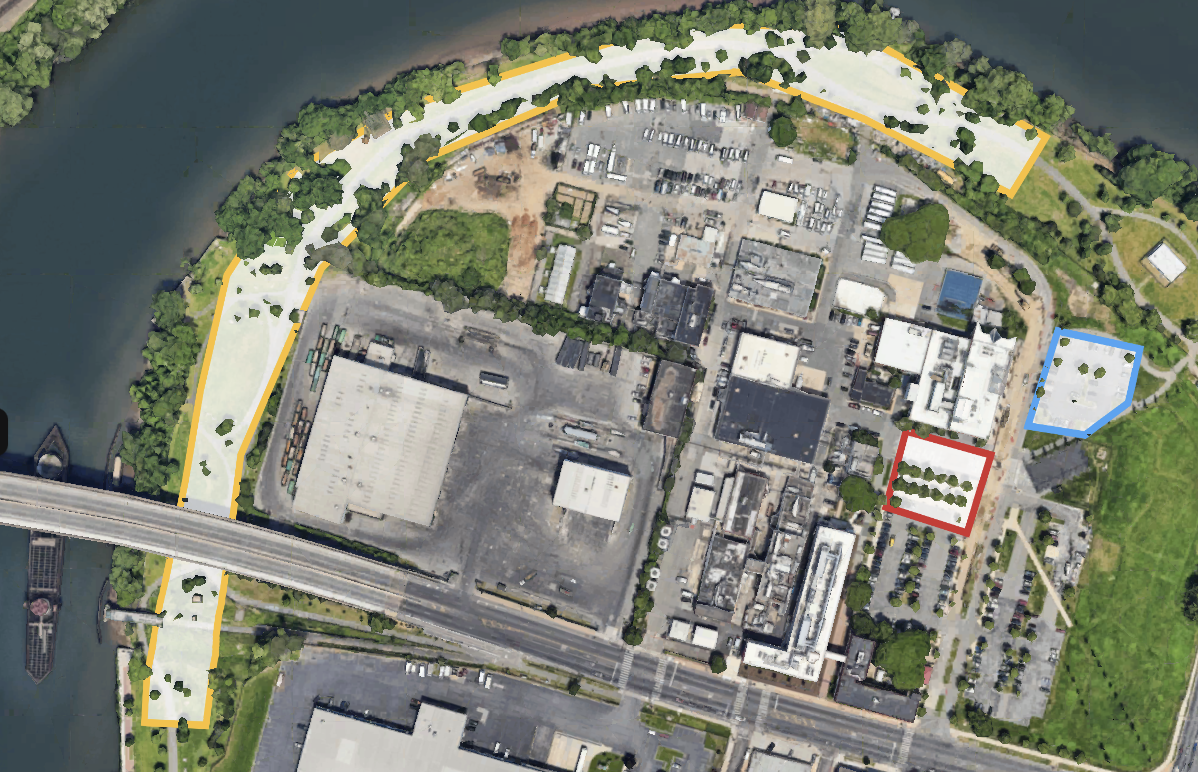}
    \caption{The outdoor experiment areas for our experiments.
    Our park experiments are in the highlighted yellow areas.
    The parking lots are in red and blue.}
    \label{fig:experiment-areas}
\end{figure}

We carried out experiments covering more than 2 \unit{\kilo\m} distance and 47,000 \unit{\square\m} of area. The experiments were conducted in an urban office complex with office space, parking lots and an outdoor park shown in \cref{fig:experiment-areas}. First, we measure the ability of our framework to load a pre-built map, localize and then navigate to the target. Then, we demonstrate that our method can be applied to completely unknown environments, incrementally construct the map and follow the utility signal to complete the task.

We compute the path length (PL) of the trajectory taken by the robot in each experiment. We compare this against two other path lengths, SP and GT, which are computed as follows. After the experiment is complete and the full map is available, we compute the shortest path (SP) on the full planning graph. This ablates away any odometry error since our traveled path and the planning graph are built on the same set of odometry measurements. To obtain the length of the optimal path (GT) from the starting position to the goal in the outdoor experiments, we measure the path length from GPS coordinates annotated manually using Google Earth. For indoor experiments, we measure the shortest path from the start to goal with a rangefinder. We measure the competitive ratio in \cref{tab:real_expts} by comparing the distance measured by the visual odometry on our robot and these privileged distance measurements, given by SP/PL and GT/PL, respectively. To show the scale of our experiments, we measure and report the approximate area of the operational area of the environment for each experiment from Google Earth, and the number of Gaussians in the resulting constructed map.

\begin{figure}[!tpb]
    \centering
    \includegraphics[width=\linewidth]{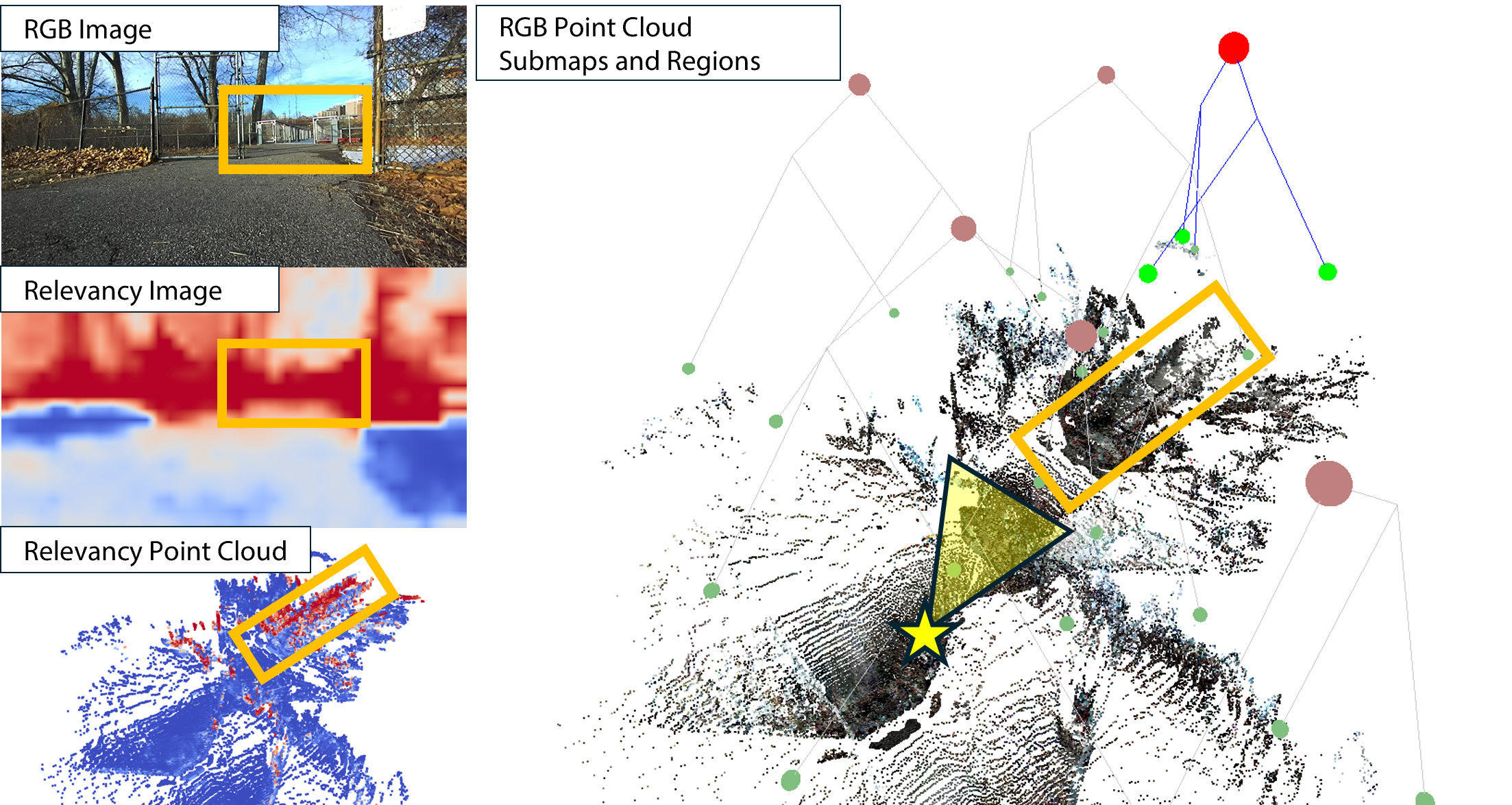}
    \caption{A visualization of the constructed dense map and associated submap and region nodes. The robot is given the task ``navigate to entrance to pier". The images show the robot's view upon completion of the task. The point clouds show the area from above at an angle. The RGB image, relevancy image and relevancy point cloud are shown on the left. The colored point cloud is shown on the right with submap (red) and region (green) nodes overlaid. The submap and region nodes that exceed the relevancy threshold are highlighted. The robot position is marked by the yellow star with its corresponding field-of-view. The pier is annotated by orange boxes in the images and point clouds. Task relevancy is colored blue to red in increasing relevance.
    }
    \label{fig:pier_viz}
\end{figure}

\subsubsection{Navigation in a pre-built map}
To show how our sparse hierarchical map can be used for navigation, we conduct an experiment (Outdoor1) where the robot uses a pre-built map to identify and plan a path to a region of interest. We first tele-operate the robot to a pier several hundred meters from the starting position and save the constructed map. On a separate run, the map is loaded on the robot. The robot is given the task ``navigate to entrance of the pier''. Task relevancy is computed across the submaps and regions, and the submap with the highest utility is identified. 
The robot plans a path to the submap of interest and navigates to the goal. \cref{tab:real_expts} shows the quantitative results for this experiment. Our method performs close to the ground truth since it has access to a pre-built map of the environment. A visualization of the map query is shown in \cref{fig:pier_viz}.

\begin{figure}[!tpb]
    \centering
    \includegraphics[width=\linewidth]{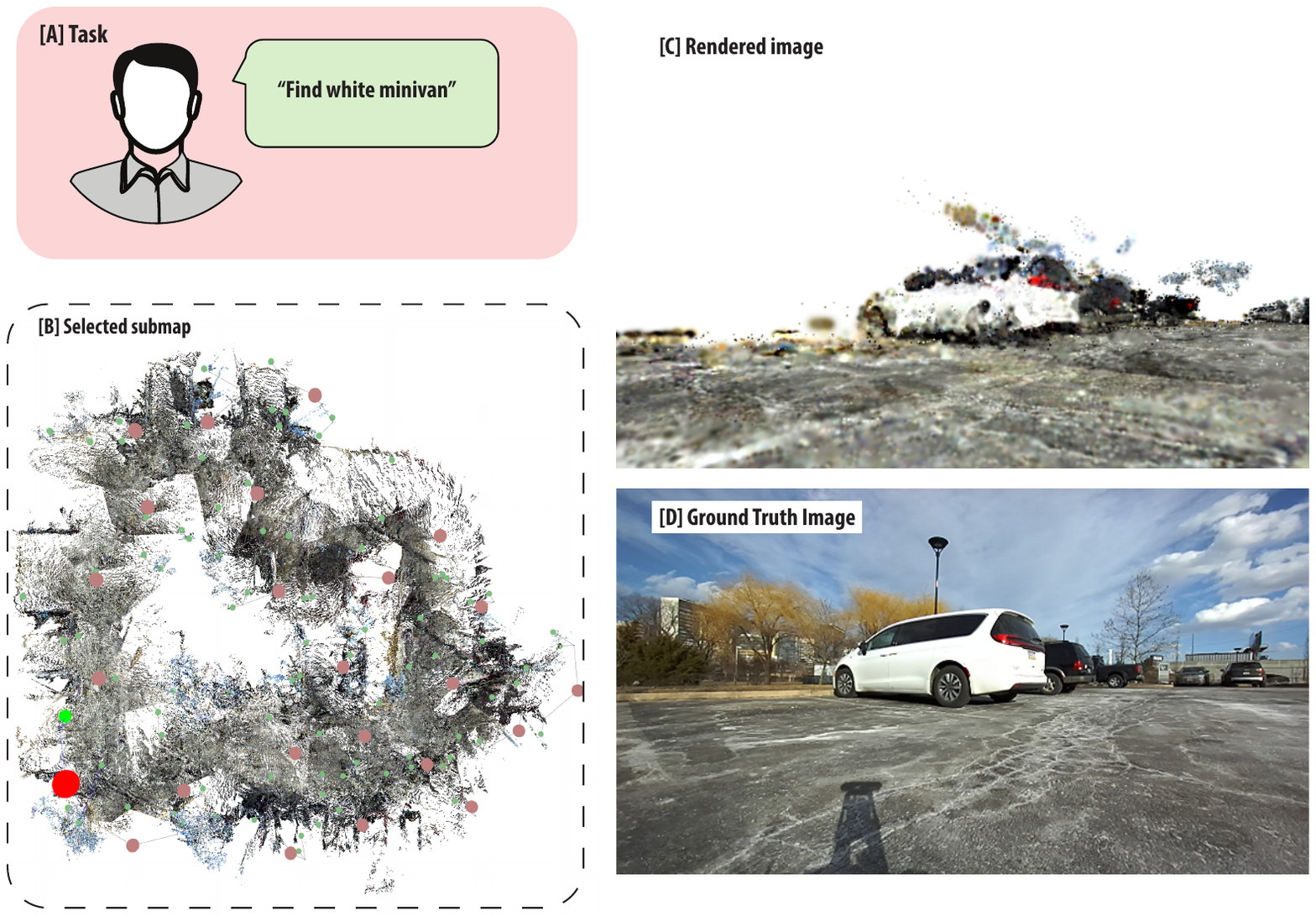}
    \caption{[A] shows the prompt for the task ``Find white minivan'' provided to ATLAS after it has built a map. [B] shows the selected submap and region in the bottom-left with highest relevance. [C] shows the rendered image from the vantage point with the highest relevance to the task. [D] shows the ground truth image.}
    \label{fig:rendering-exp}
\end{figure}

One interesting aspect of ATLAS is that we have the ability to query areas of relevance in the language-guided map and provide visual feedback by rendering images of relevant areas. Given a task query, the most relevant submap and region can be retrieved and one can render an image at a chosen viewpoint. \cref{fig:rendering-exp} shows a qualitative result of such retrieval, constructed from data collected by tele-operating our robot. While not investigated in this work, this approach enables many possibilities in providing visual feedback to the user and incorporating task clarification and re-specification during exploration and navigation.

\subsubsection{Navigation with no prior map}
We conduct six real-world experiments on exploration without a prior map---two indoor and four outdoor. In these experiments, the robot starts off with no prior map or information of the environment. Given a user-specified task, the robot proceeds to incrementally build a metric-semantic map of the environment and uses relevant information in the map to complete the task.

In Indoor1, we show that our approach is able to leverage the semantic relationships between objects in the environment to complete tasks.
With the task ``find cushions'', the robot is able to identify chairs and couches as areas of high relevance to the task and proceeds to inspect them, eventually successfully locating the cushions. \cref{fig:cushions_viz} provides a qualitative result from this experiment.
For Indoor2, we perform an object search of ``find plants'' and then demonstrate re-tasking the robot to ``exit the building''. 

\begin{figure}[!tpb]
    \centering
    \includegraphics[width=0.95\linewidth]{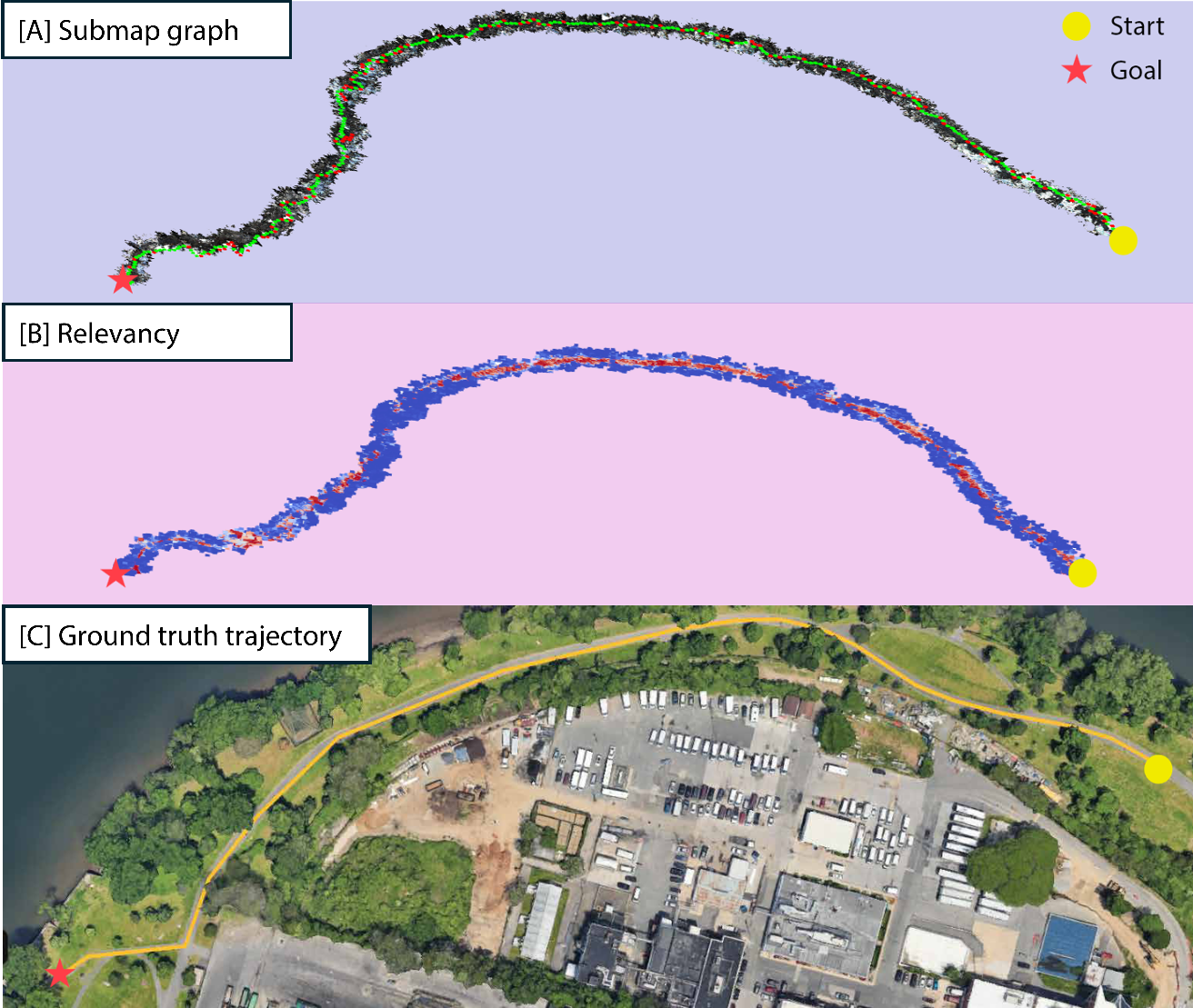}
    \caption{Top: The map built from the task ``Find boardwalk near road''. The colored Gaussians and the submap anchors (red) are visualized. The resulting hierarchical graph can then be used to retrieve and navigate in the map. The shortest path (SP) to the boardwalk is shown with the green nodes. Middle: The task relevancy of the Gaussians is colored blue (low) to red (high). Bottom: The ground truth trajectory (GT) is overlaid in yellow on an image from Google Earth to highlight the scale of the experiments.}
    \label{fig:outdoor-exp-scale-gt}
\end{figure}

For the outdoor experiments, we consider three settings. In Outdoor 2, we use an under-specified task ``inspect road blockage'' and the robot leverages semantics of the pavement to search for and identify obstructions. In Outdoor 3, the robot is tasked with ``find parking lot'' while starting close to the entrance of a building. In Outdoor 4 and Outdoor 5, we conduct large-scale experiments in the park. A visualization of task relevancy and the ground truth trajectory is shown in \cref{fig:outdoor-exp-scale-gt}.

\cref{tab:real_expts} presents a quantitative overview of these experiments. The competitive ratio of our method is $\sim$ 0.59 on SP and $\sim$ 0.56 in terms of GT. This shows that our tasks require some exploration but, in general, ATLAS performs at least half as well as a privileged baseline. In our experiments, our method is also able to store and manage more than ten million Gaussians on-board the robot.

The scale of the environment and the duration of the experiment on real robots in this paper are quite large.
Hardware and logistical complexities (battery life, reliability of the hardware, calibration issues, weather, pedestrian and vehicular traffic etc.) make it difficult to run multiple replicates of such long-range outdoor experiments, and so each real-robot experiment is run only once.
Closed-loop simulation experiments are challenging due to the lack of large-scale simulation environments; typical semantic navigation benchmarks only consider indoor navigation (e.g. HM3D).
We recognize that this methodology is inadequate to evaluate statistical repeatability.
It is however a good ``demonstration'' of the operational scope that ATLAS can handle. Our experiment is quite long and representative of the diverse situations that an autonomous robot may encounter outdoors and it thus provides a meaningful basis for evaluating our approach.
In the following section, we show the performance of different components of the system on established benchmark datasets, which allows the reader to gauge the statistical significance of the results.

\begin{figure*}[!ht]
    \centering
    \includegraphics[width=0.99\linewidth]{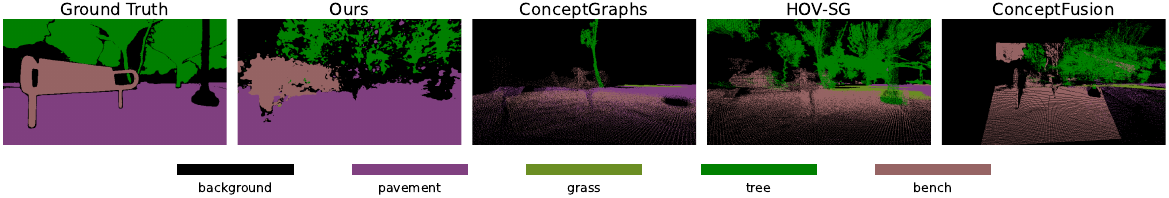}
    \caption{Dense semantic segmentation in an outdoor scene. In this example, all the baselines over-segment the bench to include the pavement. HOV-SG and ConceptGraphs use image segmentation priors to create object-level representations with sparse feature vectors. ConceptGraphs fails to segment many trees and omits them from the map entirely. ConceptFusion maintains dense features in the point cloud but also relies on image segmentation priors to extract region-level semantic features, which can also result in inaccurate semantics. In contrast, our dense representation more accurately captures both the geometry and semantics of such unstructured environments.}
    \label{fig:seg_comparison}
\end{figure*}

\section{Ablation Study}
\label{sec:seg-experiments}

In this section, we discuss ablation experiments in simulated environments on (i) open-vocabulary semantic segmentation, (ii) submapping loop closure, and (iii) memory and computational efficiency of different parts of the ATLAS pipeline. We evaluate each component of our method with prior work to obtain a detailed understanding of the performance of ATLAS in different settings and how our choices are critical for real-time robot autonomy.

\subsubsection{Open-vocabulary semantic segmentation}
We evaluate the open-vocabulary semantic segmentation performance of our mapping approach against other open-vocabulary mapping methods -- ConceptGraphs~\cite{conceptgraphs}, HOV-SG~\cite{werby2024hierarchical} and ConceptFusion~\cite{jatavallabhula2023conceptfusion}. We first compare the methods on 3D object-based semantic segmentation on indoor data. While navigation of structured indoor environments benefits from object-based semantic segmentation, outdoor environments are much less structured and lack distinct semantic objects. To evaluate performance in such unstructured environments, we also evaluate the methods on dense semantic segmentation on an outdoor scene. For the indoor dataset, we use ScanNet scenes: scene0011\textunderscore00, scene0050\textunderscore00, scene0231\textunderscore00, scene0378\textunderscore00, scene0518\textunderscore00 following the evaluation protocol used in~\cite{werby2024hierarchical}. For the outdoor scene, we collect data by tele-operating our robot and sample 500 images for reconstruction.


\medskip \textbf{Indoor experiments.} For the indoor segmentation task, we compute the 3D mean Intersection Over Union (mIOU),  mean accuracy (mAcc) and frequency-weighted mIOU (F-mIOU) following the same procedure in \cite{werby2024hierarchical}.
We also compare the memory and computation performance of the various methods.
 
The indoor segmentation results are presented in \cref{table:ov_seg}.
Qualitative results are shown in \cref{fig:agglomerative}.
In such structured environments, image-based segmentation priors as used in the other methods perform well at identifying object instances, enabling storage of uncompressed semantic features in a geometrically sparse map.
Nevertheless, with our compressed language feature representation and without using image segmentation priors, we are able to achieve comparable performance to the baselines.
Our method also requires significantly less memory and is more than an order of magnitude faster.
Such compute efficiency is critical for real-time reconstruction of large-scale environments on compute-constrained robots.

\medskip \textbf{Outdoor experiments.} We evaluate the various methods on an outdoor dense segmentation task.
We evaluate the rendered segmentation masks, color and depth images from the maps constructed by each method against the ground truth images.
We evaluate the quality of the color images with the metrics Peak Signal-to-Noise Ratio (PSNR), Structural Similarity Index~\cite{1292216} (SSIM) and Learned Perceptual Image Patch Similarity~\cite{zhang2018perceptual} (LPIPS).
We compute the Root Mean-Square-Error (RMSE) of the depth images.
Due to the lack of ground truth labels for the outdoor data, we obtain ground truth segmentation masks using Grounded SAM 2~\cite{ren2024grounded}.
We compute the 2D mIOU of the segmentation masks on the classes [`background', `pavement', `grass', `tree', `bench'].

Outdoor segmentation results are presented in \cref{table:ov_seg} with qualitative results visualized in \cref{fig:seg_comparison}.
ATLAS achieves better mIOU scores on dense semantic segmentation and also provides better dense scene reconstruction with the 3DGS representation.
The baseline methods rely on image segmentation priors for creating object clusters.
ConceptGraphs and HOV-SG maintain sparse language feature vectors for each segmented object while ConceptFusion fuses region-level feature embeddings with image-level global features in the point cloud.
As shown in \cref{fig:seg_comparison}, this can cause issues with over- or under-segmentation of the scene and limits scene understanding of highly unstructured environments.
Our dense representation can accurately capture both semantics and geometry for dense scene understanding which is well-suited for unstructured environments where explicit object instances are less common.
We note that the dense segmentation results are evaluated on Grounded SAM 2 masks which may not be completely accurate.
Nonetheless, these results reflect the relative performance between the methods and highlight the limitations of sparse representations that are built on image segmentation priors.

\begin{table*}[!ht]
\centering
\begin{adjustbox}{width=\linewidth}
\renewcommand{\arraystretch}{1.15}
\begin{tabular}{c | rrr rr | rrrrr}
\toprule
& \multicolumn{5}{c|}{\textbf{Indoor 3D Segmentation (ScanNet)}}
& \multicolumn{5}{c}{\textbf{Outdoor Dense Segmentation}} \\
\cmidrule(lr){2-6} \cmidrule(lr){7-11}

\textbf{Method}
& \textbf{mIOU} $\uparrow$ & \textbf{mAcc} $\uparrow$ & \textbf{F-mIOU} $\uparrow$
& \textbf{Memory} $\downarrow$ & \textbf{Time/Frame} $\downarrow$
& \textbf{mIOU} $\uparrow$ & \textbf{PSNR} $\uparrow$
& \textbf{SSIM} $\uparrow$
& \textbf{LPIPS} $\downarrow$
& \textbf{RMSE} $\downarrow$ \\

& & & & (GB) & (s)
& & (dB) & & & (m) \\
\midrule

ConceptGraphs~\cite{conceptgraphs}
& \cellcolor{orange!40} 0.16 & 0.20 & \cellcolor{orange!40} 0.28
& 51.93 & 5.38
& 0.07 & 5.35 & 0.20 & 0.96 & 5.93 \\

HOV-SG~\cite{werby2024hierarchical}
& \cellcolor{red!40} \textbf{0.22} & \cellcolor{red!40} \textbf{0.30} & \cellcolor{red!40} \textbf{0.43}
& 46.93 & 8.58
& \cellcolor{orange!40} 0.19 & 5.36 & 0.24 & 0.95 & \cellcolor{orange!40} 2.49 \\

ConceptFusion~\cite{jatavallabhula2023conceptfusion}
& 0.11 & 0.12 & 0.21
& \cellcolor{orange!40} 42.56 & \cellcolor{orange!40} 8.24
& 0.10 & \cellcolor{orange!40} 6.87 & \cellcolor{orange!40} 0.32
& \cellcolor{orange!40} 0.86 & 4.76 \\

ATLAS (ours)
& 0.15 & \cellcolor{orange!40} 0.27 & 0.16
& \cellcolor{red!40} \textbf{26.67} & \cellcolor{red!40} \textbf{0.21}
& \cellcolor{red!40} \textbf{0.30} & \cellcolor{red!40} \textbf{20.69}
& \cellcolor{red!40} \textbf{0.80}
& \cellcolor{red!40} \textbf{0.29}
& \cellcolor{red!40} \textbf{0.33} \\

\bottomrule
\end{tabular}
\end{adjustbox}
\caption{We evaluate the map reconstruction quality of several open-vocabulary semantic mapping methods. While 3D object-based semantic segmentation is commonly used for indoor scenes, outdoor scenes are much more unstructured and contain few object instances. We instead evaluate the dense scene understanding capability of these approaches for the outdoor scene. Left: Open-vocabulary 3D semantic segmentation, memory and computation time on ScanNet. Baseline segmentation results from \cite{werby2024hierarchical}. Our method achieves comparable results while taking more than an order of magnitude less computation time, which is essential for real-time operation.
Right: Open-vocabulary dense semantic segmentation on outdoor unstructured scenes. Unlike other map representations that use image segmentation priors to create sparse semantic clusters, our mapping approach retains dense semantics and geometry for effective scene understanding in unstructured environments.}
\label{table:ov_seg}
\end{table*}

\begin{table}[!tbh]
\centering
\begin{adjustbox}{width=\linewidth}
\renewcommand{\arraystretch}{1.1}
\begin{tabular}{crrrr}
\toprule
\multirow{2}{*}{\textbf{Method}} & \textbf{PSNR} $\uparrow$ & \textbf{SSIM} $\uparrow$ & \textbf{LPIPS} $\downarrow$ & \textbf{RMSE} $\downarrow$ \\
& (dB) & & & (m) \\ \hline  
VIO & 12.62 & 0.53 & 0.48 & 0.89 \\
VIO + PGO & \cellcolor{orange!40} 14.09 & \cellcolor{orange!40} 0.54 & \cellcolor{orange!40} 0.47 & \cellcolor{orange!40} 0.85 \\
VIO + PGO + SU & \cellcolor{red!40}\textbf{14.31} & \cellcolor{red!40}\textbf{0.59} & \cellcolor{red!40}\textbf{0.44} & \cellcolor{red!40}\textbf{0.76} \\
\midrule
GT & 16.13 & 0.65 & 0.37 & 0.39 \\
\bottomrule
\end{tabular}
\end{adjustbox}
\caption{Evaluation of loop closure-aware submap pose graph update. VIO refers to visual-inertial odometry, PGO refers to pose graph optimization, SU refers to submap anchor pose updates and GT refers to the ground truth poses. Our submapping approach enables integration of loop closures on a submap level to improve map reconstruction accuracy.}
\label{table:loop}
\end{table}

\subsubsection{Loop closure with submaps}
To evaluate the effectiveness of pose correction with submaps, we collected data by tele-operating a robot in a large indoor scene, and generated maps from this set of data.
We built three maps, one with just visual-inertial odometry (VIO) pose estimates, one with the addition of pose graph optimization (PGO), and finally one with submap anchor pose updates on top of the PGO (denoted by SU).
Since retroactively updating all poses with each iteration of PGO may require significant updates to the existing Gaussian parameters which is challenging to achieve in real time, we use only the latest optimized pose at each mapping iteration.
By only updating submap anchor poses based on the PGO, we can apply corrections to the map globally while maintaining the local consistency of Gaussians within each submap.
We use LiDAR odometry~\cite{fasterlio} to obtain ground truth poses for the test set.
We compare the rendered images from each map and evaluate the rendering quality.
Finally, to present the upper bound of the Gaussian splatting approach on this dataset, we also generate a map using the test set poses.

We evaluate the PSNR, SSIM, LPIPS on the color images and RMSE on the depth images. The results are presented in \cref{table:loop}. We show that we are able to leverage PGO with our submap-based mapping framework to achieve better reconstruction quality.
Our method is able to apply pose corrections by shifting the submap anchors.

\begin{table}[!tbh]
\centering
\begin{adjustbox}{max width=\linewidth}
\renewcommand{\arraystretch}{1.1}
\begin{tabular}{c r r r}
\toprule
\multirow{2}{*}{\textbf{Method}} & \multicolumn{3}{c}{\textbf{GPU Memory (GB)} $\downarrow$} \\
& 2m submap & 5m submap & No submap \\
\midrule
Gaussian-SLAM~\cite{yugay2023gaussian} &  36.21 & $\times$ & $\times$ \\
LoopSplat~\cite{zhu2024loopsplat} & \cellcolor{orange!40}21.33 & $\times$ & $\times$ \\
ATLAS (ours) & \cellcolor{red!40}\textbf{9.68} & \cellcolor{red!40}\textbf{12.51} & \cellcolor{red!40}\textbf{16.32} \\
\bottomrule
\end{tabular}
\end{adjustbox}
\caption{Comparison of 3DGS memory performance on scene 00824 of the HM3D dataset. The symbol $\times$ indicates that the method failed due to excessive memory requirements. Existing 3DGS methods that implement submapping do not adequately handle offloading of maps from the GPU and cannot handle mapping of large-scale environments.}
\label{table:memory}
\end{table}

\subsubsection{Memory and computation}
To highlight the value of the dynamic loading of submaps for 3D Gaussian splatting on SWaP-constrained platforms, we evaluate several 3DGS approaches on mapping a large indoor environment.
We compare our method against MonoGS~\cite{matsuki2024gaussian}, RTG-SLAM~\cite{peng2024rtg}, Gaussian-SLAM~\cite{yugay2023gaussian} and LoopSplat~\cite{zhu2024loopsplat}.
We use scene 00824, which is a large indoor scene consisting of multiple rooms, from the Habitat Matterport 3D Semantics Dataset \cite{yadav2023habitat} (HM3D), following the method in \cite{werby2024hierarchical} to generate ground truth observations and poses. 
If they support submapping, methods are evaluated with different submap sizes, at 2 m and 5 m distances between submaps.
We note that Gaussian-SLAM and LoopSplat both employ submapping but only unload submaps and do not handle reloading and updating of submaps.

The results are presented in \cref{table:memory}.
MonoGS and RTG-SLAM do not implement submapping and failed due to memory requirements.
Gaussian-SLAM~\cite{yugay2023gaussian} and LoopSplat~\cite{zhu2024loopsplat} require large amounts of memory even at relatively small submap sizes.
The results highlight that many Gaussian splatting SLAM methods cannot support memory-efficient mapping of large-scale environments, even on the scale of indoor environments.
Ensuring efficient scaling of memory is crucial for operation on robots with limited compute.

\begin{figure}
\centering
  \begin{tikzpicture}[scale=0.9,trim axis left, trim axis right]
    \pgfmathsetlengthmacro\MajorTickLength{
      \pgfkeysvalueof{/pgfplots/major tick length} * 0.05
    }
    \pgfplotsset{every tick label/.append style={font=\footnotesize}}
    \begin{axis}[ylabel={Time (ms)},
                 xlabel={Number of Gaussians},
                 xmode=log,
                 ymode=log,
                 major tick length=\MajorTickLength,
                 grid=both,
                 grid style={line width=.1pt, draw=gray!10},
                 major grid style={line width=.2pt,draw=gray!50},
                 every tick/.style={
                    black,
                    semithick
      }]

    \addplot table[x index=0,y index=1,col sep=comma] {collision-test-time.dat};
    \addplot+[forget plot,only marks] 
    plot[error bars/.cd, y dir=both, y explicit]
    table[x index=0, y index=1, y error index=2, col sep=comma] {collision-test-time.dat}; 
    \end{axis}
    
    \end{tikzpicture}
\vspace{-0.8cm}
\caption{Computation time for testing trajectory collisions on the GPU over a 5 m horizon (140 points) as a function of the number of Gaussians. We observe that the computation time increases linearly with the number of Gaussians. This emphasizes the importance of considering only a subset of Gaussians in the map for collision checking since there are typically millions of Gaussians in the map when operating in large environments.}
\label{fig:collision_checking}
\end{figure}

We also highlight the importance of submapping and choosing an appropriate local map radius for real-time trajectory collision checking.
We measure the time taken for collision checks across a 5 m planning horizon against different numbers of Gaussians.
As shown in Fig.~\ref{fig:collision_checking}, once the Gaussian parameters no longer fit within the GPU cache, computation time increases linearly with the number of Gaussians.
When operating in large-scale environments, the total number of Gaussians in our maps often reaches the order of millions.
With 10 million Gaussians in the map, as in one of our experiments, collision checking would require more than a second.
In such cases, even GPU-accelerated collision checking can become computationally prohibitive for real-time performance unless the number of Gaussians in the local map is carefully managed.

\section{Limitations and Lessons Learned}

\subsection{3DGS as a semantic map representation}
It is challenging to build a dense and semantically rich map online that can be immediately used for semantic navigation over a large scale, which led us to consider the components in our proposed method.
Distinct object-level labels are sufficient when tasks are known \emph{a priori}.
In a setting where the task is respecified or ambiguously defined, a reusable map with continuous labels is necessary.
Voxel-based mapping approaches do not maintain dense, view-dependent photometric information and cannot render complex scenes.
They are limited in terms of what downstream tasks can be supported, compared to radiance fields.
A representation built upon language-embedded 3DGS has proven to be an efficient representation for this task.
The components of such a system need to be tightly coupled to enable mapping and planning online despite compute constraints.
For instance, global bundle adjustment to optimize the entire map as done in map-then-plan approaches would work better, but it would be too expensive to run online.
To address complex semantic tasks with modern ML methods, we need to engineer efficient strategies such as incremental mapping, feature compression, and submap-level pose updates.
These are important steps towards effectively utilizing radiance fields for real-time autonomous mapping and navigation.

\subsection{Challenges of real-world autonomy}
Since we are addressing the challenges of real-world autonomy, the method is limited by the quality of sensor measurements.
One such limitation is a heavy dependence on the quality of the external odometry solution. 
Visual-inertial odometry can perform poorly in outdoor conditions such as harsh sunlight, textureless regions on the road and grass, and long-term drift due to the lack of distinct features in the environment.
As shown in \cref{table:loop}, the map constructed from VIO state estimates is of lower quality compared to one built with LiDAR-inertial odometry estimates.
Likewise, vision-based depth estimation can be noisy, especially over textureless regions and at edges of objects.
These errors can significantly degrade the visual quality of reconstructions, evident in the noisy rendering of the car in Fig.~\ref{fig:rendering-exp}, where erroneous measurements of the road and trees are blended in with the edges of the vehicle.
In addition, while we have shown that it is possible to efficiently incorporate loop closures into the 3DGS map, the system is fundamentally reliant on the external state estimation module’s ability to accurately detect and handle these loop closures.
These errors can severely impact the system's ability to generate accurate task-driven semantics, as these factors form the core of the data used to interpret and solve tasks.
\subsection{Vision-language features for task-driven navigation}
While there is substantial work in using learning-based autoencoders for compressing feature vectors, we have found empirically that a linear dimensionality reduction technique like PCA retains sufficient semantic information for many tasks.
In scenarios that demand efficient and generalizable compression, PCA is a simple yet effective approach.
This method faces challenges when dealing with very abstract tasks. While it performs well for tasks with concrete and well-defined semantics, it may struggle with tasks that require a deeper understanding or decomposition of more abstract concepts.
For instance, specific tasks such as ``find a bench'' are well-defined in semantics, whereas a more abstract task like ``find a place to rest'' may require further reasoning.
In such cases, additional tools, such as large language models (LLMs), may be necessary to decompose the tasks into more manageable components but potentially adding another layer of complexity.

\section{Conclusion} 
\label{sec:conclusion}

In this paper, we develop a framework and methodology that allow robots to autonomously explore and navigate unstructured environments to accomplish tasks that are specified using natural language. 
Central to our approach is a hierarchical representation built on language-embedded Gaussian splatting that can be run on-board a robot in real time.
We use indoor and outdoor experiments involving traversals of hundreds of meters to show the robustness of our approach and our ability to build large-scale maps and identify semantically relevant objects in diverse environments.
Our method does not rely on the presence of structure or rich semantics in the scene and can be applied in general settings.
In future work, we plan to incorporate this system with a higher-level planning framework such as an LLM-based planner for task decomposition to enable more abstract reasoning of tasks.
While this paper empirically evaluates this method on largely planar environments, in future work we will consider multi-floor buildings and large environments with elevation changes.

\bibliographystyle{unsrt}
\bibliography{references}

@string{IROS        = "IEEE/RSJ Int.~Conf.~on Intelligent Robots and Systems"}

@string{ICRA        = "IEEE Int.~Conf.~on Robotics \& Automation"}

@string{RAL         = "IEEE Robotics and Automation Letters"}

@string{RSS         = "Proc.~of Robotics: Science and Systems"}

@string{ICCV        = "IEEE/CVF Int.~Conf.~on Computer Vision"}

@string{CVPR        = "Proc.~of IEEE Conf.~on  Computer Vision and Pattern Recognition"}

@string{TRO         = "IEEE Transactions on Robotics"}

@InProceedings{Ranzinger_2024_CVPR,
    author    = {Ranzinger, Mike and Heinrich, Greg and Kautz, Jan and Molchanov, Pavlo},
    title     = {AM-RADIO: Agglomerative Vision Foundation Model Reduce All Domains Into One},
    booktitle = {Proceedings of the IEEE/CVF Conference on Computer Vision and Pattern Recognition (CVPR)},
    month     = {June},
    year      = {2024},
    pages     = {12490-12500}
}

@inproceedings{steinke2025collaborative,
  title={Collaborative Dynamic 3D Scene Graphs for Open-Vocabulary Urban Scene Understanding},
  author={Steinke, Tim and B{\"u}chner, Martin and V{\"o}disch, Niclas and Valada, Abhinav},
  booktitle={2025 IEEE/RSJ International Conference on Intelligent Robots and Systems (IROS)},
  pages={6000--6007},
  year={2025},
  organization={IEEE}
}

@article{bavle2023s,
  title={S-graphs+: Real-time localization and mapping leveraging hierarchical representations},
  author={Bavle, Hriday and Sanchez-Lopez, Jose Luis and Shaheer, Muhammad and Civera, Javier and Voos, Holger},
  journal=RAL,
  volume={8},
  number={8},
  pages={4927--4934},
  year={2023},
  publisher={IEEE}
}

@inproceedings{looper20233d,
  title={3d vsg: Long-term semantic scene change prediction through 3d variable scene graphs},
  author={Looper, Samuel and Rodriguez-Puigvert, Javier and Siegwart, Roland and Cadena, Cesar and Schmid, Lukas},
  booktitle=ICRA,
  pages={8179--8186},
  year={2023},
  organization={IEEE}
}

@inproceedings{wu2021scenegraphfusion,
  title={Scenegraphfusion: Incremental 3d scene graph prediction from rgb-d sequences},
  author={Wu, Shun-Cheng and Wald, Johanna and Tateno, Keisuke and Navab, Nassir and Tombari, Federico},
  booktitle=CVPR,
  pages={7515--7525},
  year={2021}
}

@inproceedings{armeni20193d,
  title={3d scene graph: A structure for unified semantics, 3d space, and camera},
  author={Armeni, Iro and He, Zhi-Yang and Gwak, JunYoung and Zamir, Amir R and Fischer, Martin and Malik, Jitendra and Savarese, Silvio},
  booktitle=ICCV,
  pages={5664--5673},
  year={2019}
}

@ARTICLE{maggio2024clio,
  author={Maggio, Dominic and Chang, Yun and Hughes, Nathan and Trang, Matthew and Griffith, Dan and Dougherty, Carlyn and Cristofalo, Eric and Schmid, Lukas and Carlone, Luca},
  journal={IEEE Robotics and Automation Letters}, 
  title={Clio: Real-Time Task-Driven Open-Set 3D Scene Graphs}, 
  year={2024},
  volume={9},
  number={10},
  pages={8921-8928},
  doi={10.1109/LRA.2024.3451395}}

@article{asgharivaskasi2023semantic,
  title={Semantic OcTree mapping and Shannon mutual information computation for robot exploration},
  author={Asgharivaskasi, Arash and Atanasov, Nikolay},
  journal=TRO,
  volume={39},
  number={3},
  pages={1910--1928},
  year={2023},
  publisher={IEEE}
}

@inproceedings{papatheodorou2023finding,
  title={Finding Things in the Unknown: Semantic Object-Centric Exploration with an MAV},
  author={Papatheodorou, Sotiris and Funk, Nils and Tzoumanikas, Dimos and Choi, Christopher and Xu, Binbin and Leutenegger, Stefan},
  booktitle={2023 IEEE International Conference on Robotics and Automation (ICRA)},
  pages={3339--3345},
  year={2023},
  organization={IEEE}
}

@article{schmid2021unified,
  title={A unified approach for autonomous volumetric exploration of large scale environments under severe odometry drift},
  author={Schmid, Lukas and Reijgwart, Victor and Ott, Lionel and Nieto, Juan and Siegwart, Roland and Cadena, Cesar},
  journal={IEEE Robotics and Automation Letters},
  volume={6},
  number={3},
  pages={4504--4511},
  year={2021},
  publisher={IEEE}
}

@INPROCEEDINGS{bai2016bayesianexp,
  author={Bai, Shi and Wang, Jinkun and Chen, Fanfei and Englot, Brendan},
  booktitle={2016 IEEE/RSJ International Conference on Intelligent Robots and Systems (IROS)}, 
  title={Information-theoretic exploration with Bayesian optimization}, 
  year={2016},
  volume={},
  number={},
  pages={1816-1822},
  doi={10.1109/IROS.2016.7759289}}

@article{hughes2024ijrr,
author = {Nathan Hughes and Yun Chang and Siyi Hu and Rajat Talak and Rumaia Abdulhai and Jared Strader and Luca Carlone},
title ={Foundations of spatial perception for robotics: Hierarchical representations and real-time systems},
journal = {The International Journal of Robotics Research},
year = {2024},
doi = {10.1177/02783649241229725},
URL = {https://doi.org/10.1177/02783649241229725},
eprint = {https://doi.org/10.1177/02783649241229725}
}

@article{Strader24ral-autoAbstr,
  author={Strader, Jared and Hughes, Nathan and Chen, William and Speranzon, Alberto and Carlone, Luca},
  journal={IEEE Robotics and Automation Letters}, 
  title={Indoor and Outdoor 3D Scene Graph Generation Via Language-Enabled Spatial Ontologies}, 
  year={2024},
  volume={9},
  number={6},
  pages={4886-4893},
  doi={10.1109/LRA.2024.3384084}
}

@article{tao2024rt,
  title={Rt-guide: Real-time gaussian splatting for information-driven exploration},
  author={Tao, Yuezhan and Ong, Dexter and Murali, Varun and Spasojevic, Igor and Chaudhari, Pratik and Kumar, Vijay},
  journal={IEEE Robotics and Automation Letters},
  year={2025},
  publisher={IEEE}
}

@inproceedings{dang2018autonomous,
  title={Autonomous exploration and simultaneous object search using aerial robots},
  author={Dang, Tung and Papachristos, Christos and Alexis, Kostas},
  booktitle={2018 IEEE Aerospace Conference},
  pages={1--7},
  year={2018},
  organization={IEEE}
}

@inproceedings{yokoyama2024vlfm,
  title={Vlfm: Vision-language frontier maps for zero-shot semantic navigation},
  author={Yokoyama, Naoki and Ha, Sehoon and Batra, Dhruv and Wang, Jiuguang and Bucher, Bernadette},
  booktitle={2024 IEEE International Conference on Robotics and Automation (ICRA)},
  pages={42--48},
  year={2024},
  organization={IEEE}
}

@inproceedings{bircher2016receding,
  title={Receding horizon" next-best-view" planner for 3d exploration},
  author={Bircher, Andreas and Kamel, Mina and Alexis, Kostas and Oleynikova, Helen and Siegwart, Roland},
  booktitle={2016 IEEE international conference on robotics and automation (ICRA)},
  pages={1462--1468},
  year={2016},
  organization={IEEE}
}

@inproceedings{charrow2015information,
  title={Information-theoretic mapping using cauchy-schwarz quadratic mutual information},
  author={Charrow, Benjamin and Liu, Sikang and Kumar, Vijay and Michael, Nathan},
  booktitle={2015 IEEE International Conference on Robotics and Automation (ICRA)},
  pages={4791--4798},
  year={2015},
  organization={IEEE}
}

@online{iot4ag,
    author = "",
    title = "{The Internet of Things for Precision Agriculture (IoT4Ag), an NSF Engineering Research Center}",
    url  = "https://iot4ag.us"
}

@INPROCEEDINGS{werby2024hierarchical, 
    AUTHOR    = {Abdelrhman Werby AND Chenguang Huang AND Martin Büchner AND Abhinav Valada AND Wolfram Burgard}, 
    TITLE     = {{Hierarchical Open-Vocabulary 3D Scene Graphs for Language-Grounded Robot Navigation}}, 
    BOOKTITLE = {Proceedings of Robotics: Science and Systems}, 
    YEAR      = {2024}, 
    ADDRESS   = {Delft, Netherlands}, 
    MONTH     = {July}, 
    DOI       = {10.15607/RSS.2024.XX.077} 
}

@inproceedings{ravichandran2024spine,
  title={Spine: Online semantic planning for missions with incomplete natural language specifications in unstructured environments},
  author={Ravichandran, Zachary and Murali, Varun and Tzes, Mariliza and Pappas, George J and Kumar, Vijay},
  booktitle={2025 IEEE International Conference on Robotics and Automation (ICRA)},
  pages={13714--13721},
  year={2025},
  organization={IEEE}
}

@article{ray2024task,
  title={Task and Motion Planning in Hierarchical 3D Scene Graphs},
  author={Ray, Aaron and Bradley, Christopher and Carlone, Luca and Roy, Nicholas},
  journal={arXiv preprint arXiv:2403.08094},
  year={2024}
}

@inproceedings{shi2024language,
  title={Language embedded 3d gaussians for open-vocabulary scene understanding},
  author={Shi, Jin-Chuan and Wang, Miao and Duan, Hao-Bin and Guan, Shao-Hua},
  booktitle={Proceedings of the IEEE/CVF Conference on Computer Vision and Pattern Recognition},
  pages={5333--5343},
  year={2024}
}

@inproceedings{hu2025cg,
  title={Cg-slam: Efficient dense rgb-d slam in a consistent uncertainty-aware 3d gaussian field},
  author={Hu, Jiarui and Chen, Xianhao and Feng, Boyin and Li, Guanglin and Yang, Liangjing and Bao, Hujun and Zhang, Guofeng and Cui, Zhaopeng},
  booktitle={European Conference on Computer Vision},
  pages={93--112},
  year={2025},
  organization={Springer}
}

@inproceedings{yuezhantao2023seer,
  title={{SEER}: Safe efficient exploration for aerial robots using learning to predict information gain},
  author={Tao, Yuezhan and Wu, Yuwei and Li, Beiming and Cladera, Fernando and Zhou, Alex and Thakur, Dinesh and Kumar, Vijay},
  booktitle={2023 IEEE International Conference on Robotics and Automation (ICRA)},
  pages={1235--1241},
  year={2023},
  organization={IEEE}
}

@article{LukasIG,
  title={An efficient sampling-based method for online informative path planning in unknown environments},
  author={Schmid, Lukas and Pantic, Michael and Khanna, Raghav and Ott, Lionel and Siegwart, Roland and Nieto, Juan},
  journal={IEEE Robotics and Automation Letters},
  volume={5},
  number={2},
  pages={1500--1507},
  year={2020},
  publisher={IEEE}
}

@inproceedings{saulnier2020information,
  title={Information theoretic active exploration in signed distance fields},
  author={Saulnier, Kelsey and Atanasov, Nikolay and Pappas, George J and Kumar, Vijay},
  booktitle={2020 IEEE International Conference on Robotics and Automation (ICRA)},
  pages={4080--4085},
  year={2020},
  organization={IEEE}
}

@INPROCEEDINGS{alexis2020MP,
  author={Dharmadhikari, Mihir and Dang, Tung and Solanka, Lukas and Loje, Johannes and Nguyen, Huan and Khedekar, Nikhil and Alexis, Kostas},
  booktitle={2020 IEEE International Conference on Robotics and Automation (ICRA)}, 
  title={Motion Primitives-based Path Planning for Fast and Agile Exploration using Aerial Robots}, 
  year={2020},
  volume={},
  number={},
  pages={179-185},
  doi={10.1109/ICRA40945.2020.9196964}
}

@INPROCEEDINGS{Papachristos17UncertaintyIG,
  author={Papachristos, Christos and Khattak, Shehryar and Alexis, Kostas},
  booktitle={2017 IEEE International Conference on Robotics and Automation (ICRA)}, 
  title={Uncertainty-aware receding horizon exploration and mapping using aerial robots}, 
  year={2017},
  volume={},
  number={},
  pages={4568-4575},
  doi={10.1109/ICRA.2017.7989531}}

@ARTICLE{ASPP,
  author={Chen, Liang-Chieh and Papandreou, George and Kokkinos, Iasonas and Murphy, Kevin and Yuille, Alan L.},
  journal={IEEE Transactions on Pattern Analysis and Machine Intelligence}, 
  title={DeepLab: Semantic Image Segmentation with Deep Convolutional Nets, Atrous Convolution, and Fully Connected CRFs}, 
  year={2018},
  volume={40},
  number={4},
  pages={834-848},
  doi={10.1109/TPAMI.2017.2699184}}

@ARTICLE{tao-24-3dactivemsslam,
  author={Tao, Yuezhan and Liu, Xu and Spasojevic, Igor and Agarwal, Saurav and Kumar, Vijay},
  journal={IEEE Robotics and Automation Letters}, 
  title={3D Active Metric-Semantic SLAM}, 
  year={2024},
  volume={},
  number={},
  pages={1-8},
  doi={10.1109/LRA.2024.3363542}}

@article{devarakonda2024orionnav,
  title={OrionNav: Online Planning for Robot Autonomy with Context-Aware LLM and Open-Vocabulary Semantic Scene Graphs},
  author={Devarakonda, Venkata Naren and Goswami, Raktim Gautam and Kaypak, Ali Umut and Patel, Naman and Khorrambakht, Rooholla and Krishnamurthy, Prashanth and Khorrami, Farshad},
  journal={arXiv preprint arXiv:2410.06239},
  year={2024}
}

@inproceedings{ravichandran2022hierarchical,
  title={Hierarchical representations and explicit memory: Learning effective navigation policies on 3d scene graphs using graph neural networks},
  author={Ravichandran, Zachary and Peng, Lisa and Hughes, Nathan and Griffith, J Daniel and Carlone, Luca},
  booktitle={2022 International Conference on Robotics and Automation (ICRA)},
  pages={9272--9279},
  year={2022},
  organization={IEEE}
}

@article{amiri2022reasoning,
  title={Reasoning with scene graphs for robot planning under partial observability},
  author={Amiri, Saeid and Chandan, Kishan and Zhang, Shiqi},
  journal={IEEE Robotics and Automation Letters},
  volume={7},
  number={2},
  pages={5560--5567},
  year={2022},
  publisher={IEEE}
}

@article{kerbl20233dgs,
  title={3d gaussian splatting for real-time radiance field rendering},
  author={Kerbl, Bernhard and Kopanas, Georgios and Leimk{\"u}hler, Thomas and Drettakis, George},
  journal={ACM Transactions on Graphics},
  volume={42},
  number={4},
  pages={1--14},
  year={2023},
  publisher={ACM}
}

@inproceedings{keetha2024splatam,
        title={SplaTAM: Splat, Track \& Map 3D Gaussians for Dense RGB-D SLAM},
        author={Keetha, Nikhil and Karhade, Jay and Jatavallabhula, Krishna Murthy and Yang, Gengshan and Scherer, Sebastian and Ramanan, Deva and Luiten, Jonathon},
        booktitle={Proceedings of the IEEE/CVF Conference on Computer Vision and Pattern Recognition},
        year={2024}
}

@inproceedings{lin2014microsoft,
  title={Microsoft coco: Common objects in context},
  author={Lin, Tsung-Yi and Maire, Michael and Belongie, Serge and Hays, James and Perona, Pietro and Ramanan, Deva and Doll{\'a}r, Piotr and Zitnick, C Lawrence},
  booktitle={Computer Vision--ECCV 2014: 13th European Conference, Zurich, Switzerland, September 6-12, 2014, Proceedings, Part V 13},
  pages={740--755},
  year={2014},
  organization={Springer}
}

@inproceedings{zhu2024loopsplat,
  title={Loopsplat: Loop closure by registering 3d gaussian splats},
  author={Zhu, Liyuan and Li, Yue and Sandstr{\"o}m, Erik and Huang, Shengyu and Schindler, Konrad and Armeni, Iro},
  booktitle={2025 International Conference on 3D Vision (3DV)},
  pages={156--167},
  year={2025},
  organization={IEEE}
}

@inproceedings{qin2024langsplat,
  title={Langsplat: 3d language gaussian splatting},
  author={Qin, Minghan and Li, Wanhua and Zhou, Jiawei and Wang, Haoqian and Pfister, Hanspeter},
  booktitle={Proceedings of the IEEE/CVF Conference on Computer Vision and Pattern Recognition},
  pages={20051--20060},
  year={2024}
}

@inproceedings{matsuki2024gaussian,
  title={Gaussian splatting slam},
  author={Matsuki, Hidenobu and Murai, Riku and Kelly, Paul HJ and Davison, Andrew J},
  booktitle={Proceedings of the IEEE/CVF Conference on Computer Vision and Pattern Recognition},
  pages={18039--18048},
  year={2024}
}

@INPROCEEDINGS{he2023active,
  author={Siming, H. and Hsu, Christopher D. and Ong, Dexter and Shao, Yifei Simon and Chaudhari, Pratik},
  booktitle={2024 American Control Conference (ACC)}, 
  title={Active Perception using Neural Radiance Fields}, 
  year={2024},
  volume={},
  number={},
  pages={4353-4358},
  doi={10.23919/ACC60939.2024.10645027}}

@inproceedings{liu2017mpl,
  title={Search-based motion planning for quadrotors using linear quadratic minimum time control},
  author={Liu, Sikang and Atanasov, Nikolay and Mohta, Kartik and Kumar, Vijay},
  booktitle={2017 IEEE/RSJ international conference on intelligent robots and systems (IROS)},
  pages={2872--2879},
  year={2017},
  organization={IEEE}
}

@inproceedings{he2024active,
  title={An Active Perception Game for Robust Information Gathering},
  author={He, Siming and Tao, Yuezhan and Spasojevic, Igor and Kumar, Vijay and Chaudhari, Pratik},
  booktitle={2025 IEEE International Conference on Robotics and Automation (ICRA)},
  pages={14168--14174},
  year={2025},
  organization={IEEE}
}

@inproceedings{georgakis2021learning,
  title={Learning to Map for Active Semantic Goal Navigation},
  author={Georgakis, Georgios and Bucher, Bernadette and Schmeckpeper, Karl and Singh, Siddharth and Daniilidis, Kostas},
  booktitle={International Conference on Learning Representations}
}

@INPROCEEDINGS{1292216,
  author={Wang, Z. and Simoncelli, E.P. and Bovik, A.C.},
  booktitle={The Thrity-Seventh Asilomar Conference on Signals, Systems \& Computers, 2003}, 
  title={Multiscale structural similarity for image quality assessment}, 
  year={2003},
  volume={2},
  number={},
  pages={1398-1402 Vol.2},
  doi={10.1109/ACSSC.2003.1292216}}

@inproceedings{zhang2018perceptual,
  title={The Unreasonable Effectiveness of Deep Features as a Perceptual Metric},
  author={Zhang, Richard and Isola, Phillip and Efros, Alexei A and Shechtman, Eli and Wang, Oliver},
  booktitle={CVPR},
  year={2018}
}

@misc{ren2024grounded,
      title={Grounded SAM: Assembling Open-World Models for Diverse Visual Tasks}, 
      author={Tianhe Ren and Shilong Liu and Ailing Zeng and Jing Lin and Kunchang Li and He Cao and Jiayu Chen and Xinyu Huang and Yukang Chen and Feng Yan and Zhaoyang Zeng and Hao Zhang and Feng Li and Jie Yang and Hongyang Li and Qing Jiang and Lei Zhang},
      year={2024},
      eprint={2401.14159},
      archivePrefix={arXiv},
      primaryClass={cs.CV}
}

@inproceedings{jiang2023fisherrfactiveviewselection,
  title={Fisherrf: Active view selection and mapping with radiance fields using fisher information},
  author={Jiang, Wen and Lei, Boshu and Daniilidis, Kostas},
  booktitle={European Conference on Computer Vision},
  pages={422--440},
  year={2024},
  organization={Springer}
}

@inproceedings{xu2024hgs,
  title={Hgs-planner: Hierarchical planning framework for active scene reconstruction using 3d gaussian splatting},
  author={Xu, Zijun and Jin, Rui and Wu, Ke and Zhao, Yi and Zhang, Zhiwei and Zhao, Jieru and Gao, Fei and Gan, Zhongxue and Ding, Wenchao},
  booktitle={2025 IEEE International Conference on Robotics and Automation (ICRA)},
  pages={14161--14167},
  year={2025},
  organization={IEEE}
}

@article{hughes2022hydra,
    title={Hydra: A Real-time Spatial Perception System for {3D} Scene Graph Construction and Optimization},
    author={Nathan Hughes and Yun Chang and Luca Carlone},
    journal={Robotics: Science and Systems XVIII},
    year={2022},
}

@ARTICLE{Lee2022nerf3drecon,
  author={Lee, Soomin and Chen, Le and Wang, Jiahao and Liniger, Alexander and Kumar, Suryansh and Yu, Fisher},
  journal={IEEE Robotics and Automation Letters}, 
  title={Uncertainty Guided Policy for Active Robotic 3D Reconstruction Using Neural Radiance Fields}, 
  year={2022},
  volume={7},
  number={4},
  pages={12070-12077},
  doi={10.1109/LRA.2022.3212668}}

@article{zhan2022activermap,
  title={Activermap: Radiance field for active mapping and planning},
  author={Zhan, Huangying and Zheng, Jiyang and Xu, Yi and Reid, Ian and Rezatofighi, Hamid},
  journal={arXiv preprint arXiv:2211.12656},
  year={2022}
}

@inproceedings{feng2024naruto,
  title={Naruto: Neural active reconstruction from uncertain target observations},
  author={Feng, Ziyue and Zhan, Huangying and Chen, Zheng and Yan, Qingan and Xu, Xiangyu and Cai, Changjiang and Li, Bing and Zhu, Qilun and Xu, Yi},
  booktitle={Proceedings of the IEEE/CVF Conference on Computer Vision and Pattern Recognition},
  pages={21572--21583},
  year={2024}
}

@inproceedings{yan2023active,
  title={Active neural mapping},
  author={Yan, Zike and Yang, Haoxiang and Zha, Hongbin},
  booktitle={Proceedings of the IEEE/CVF International Conference on Computer Vision},
  pages={10981--10992},
  year={2023}
}

@article{ran2023neurar,
  title={Neurar: Neural uncertainty for autonomous 3d reconstruction with implicit neural representations},
  author={Ran, Yunlong and Zeng, Jing and He, Shibo and Chen, Jiming and Li, Lincheng and Chen, Yingfeng and Lee, Gimhee and Ye, Qi},
  journal={IEEE Robotics and Automation Letters},
  volume={8},
  number={2},
  pages={1125--1132},
  year={2023},
  publisher={IEEE}
}

@inproceedings{pan2022activenerf,
  title={Activenerf: Learning where to see with uncertainty estimation},
  author={Pan, Xuran and Lai, Zihang and Song, Shiji and Huang, Gao},
  booktitle={European Conference on Computer Vision},
  pages={230--246},
  year={2022},
  organization={Springer}
}

@misc{jin2024gsplanner,
      title={GS-Planner: A Gaussian-Splatting-based Planning Framework for Active High-Fidelity Reconstruction}, 
      author={Rui Jin and Yuman Gao and Yingjian Wang and Haojian Lu and Fei Gao},
      year={2024},
      eprint={2405.10142},
      archivePrefix={arXiv},
      primaryClass={cs.RO},
      url={https://arxiv.org/abs/2405.10142}, 
}

@article{chen2024splat,
  title={Splat-Nav: Safe Real-Time Robot Navigation in Gaussian Splatting Maps},
  author={Chen, Timothy and Shorinwa, Ola and Zeng, Weijia and Bruno, Joseph and Dames, Philip and Schwager, Mac},
  journal={arXiv preprint arXiv:2403.02751},
  year={2024}
}

@inproceedings{peng2024rtg,
  title={Rtg-slam: Real-time 3d reconstruction at scale using gaussian splatting},
  author={Peng, Zhexi and Shao, Tianjia and Liu, Yong and Zhou, Jingke and Yang, Yin and Wang, Jingdong and Zhou, Kun},
  booktitle={ACM SIGGRAPH 2024 Conference Papers},
  pages={1--11},
  year={2024}
}

@article{zuo2024fmgs,
  title={Fmgs: Foundation model embedded 3d gaussian splatting for holistic 3d scene understanding},
  author={Zuo, Xingxing and Samangouei, Pouya and Zhou, Yunwen and Di, Yan and Li, Mingyang},
  journal={International Journal of Computer Vision},
  pages={1--17},
  year={2024},
  publisher={Springer}
}

@article{yugay2023gaussian,
  title={Gaussian-slam: Photo-realistic dense slam with gaussian splatting},
  author={Yugay, Vladimir and Li, Yue and Gevers, Theo and Oswald, Martin R},
  journal={arXiv preprint arXiv:2312.10070},
  year={2023}
}

@inproceedings{wysoczanska2025clip,
  title={CLIP-DINOiser: Teaching CLIP a few DINO tricks for open-vocabulary semantic segmentation},
  author={Wysocza{\'n}ska, Monika and Sim{\'e}oni, Oriane and Ramamonjisoa, Micha{\"e}l and Bursuc, Andrei and Trzci{\'n}ski, Tomasz and P{\'e}rez, Patrick},
  booktitle={European Conference on Computer Vision},
  pages={320--337},
  year={2025},
  organization={Springer}
}

@inproceedings{dong2023maskclip,
  title={Maskclip: Masked self-distillation advances contrastive language-image pretraining},
  author={Dong, Xiaoyi and Bao, Jianmin and Zheng, Yinglin and Zhang, Ting and Chen, Dongdong and Yang, Hao and Zeng, Ming and Zhang, Weiming and Yuan, Lu and Chen, Dong and others},
  booktitle={Proceedings of the IEEE/CVF Conference on Computer Vision and Pattern Recognition},
  pages={10995--11005},
  year={2023}
}

@article{oquab2023dinov2,
  title={DINOv2: Learning Robust Visual Features without Supervision},
  author={Oquab, Maxime and Darcet, Timoth{\'e}e and Moutakanni, Th{\'e}o and Vo, Huy V and Szafraniec, Marc and Khalidov, Vasil and Fernandez, Pierre and HAZIZA, Daniel and Massa, Francisco and El-Nouby, Alaaeldin and others},
  journal={Transactions on Machine Learning Research}
}

@inproceedings{kerr2023lerf,
  title={Lerf: Language embedded radiance fields},
  author={Kerr, Justin and Kim, Chung Min and Goldberg, Ken and Kanazawa, Angjoo and Tancik, Matthew},
  booktitle={Proceedings of the IEEE/CVF International Conference on Computer Vision},
  pages={19729--19739},
  year={2023}
}

@inproceedings{yu2024language,
  title={Language-embedded gaussian splats (legs): Incrementally building room-scale representations with a mobile robot},
  author={Yu, Justin and Hari, Kush and Srinivas, Kishore and El-Refai, Karim and Rashid, Adam and Kim, Chung Min and Kerr, Justin and Cheng, Richard and Irshad, Muhammad Zubair and Balakrishna, Ashwin and others},
  booktitle={2024 IEEE/RSJ International Conference on Intelligent Robots and Systems (IROS)},
  pages={13326--13332},
  year={2024},
  organization={IEEE}
}

@article{ross2008incremental,
  title={Incremental learning for robust visual tracking},
  author={Ross, David A and Lim, Jongwoo and Lin, Ruei-Sung and Yang, Ming-Hsuan},
  journal={International journal of computer vision},
  volume={77},
  pages={125--141},
  year={2008},
  publisher={Springer}
}

@inproceedings{yadav2023habitat,
  title={Habitat-matterport 3d semantics dataset},
  author={Yadav, Karmesh and Ramrakhya, Ram and Ramakrishnan, Santhosh Kumar and Gervet, Theo and Turner, John and Gokaslan, Aaron and Maestre, Noah and Chang, Angel Xuan and Batra, Dhruv and Savva, Manolis and others},
  booktitle={Proceedings of the IEEE/CVF Conference on Computer Vision and Pattern Recognition},
  pages={4927--4936},
  year={2023}
}

@article{li2024llava,
  title={Llava-onevision: Easy visual task transfer},
  author={Li, Bo and Zhang, Yuanhan and Guo, Dong and Zhang, Renrui and Li, Feng and Zhang, Hao and Zhang, Kaichen and Zhang, Peiyuan and Li, Yanwei and Liu, Ziwei and others},
  journal={arXiv preprint arXiv:2408.03326},
  year={2024}
}

@article{lattanzi2017review,
  title={Review of robotic infrastructure inspection systems},
  author={Lattanzi, David and Miller, Gregory},
  journal={Journal of Infrastructure Systems},
  volume={23},
  number={3},
  pages={04017004},
  year={2017},
  publisher={American Society of Civil Engineers}
}

@article{chung2023into,
  title={Into the robotic depths: analysis and insights from the DARPA subterranean challenge},
  author={Chung, Timothy H and Orekhov, Viktor and Maio, Angela},
  journal={Annual Review of Control, Robotics, and Autonomous Systems},
  volume={6},
  number={1},
  pages={477--502},
  year={2023},
  publisher={Annual Reviews}
}

@article{fountas2020agricultural,
  title={Agricultural robotics for field operations},
  author={Fountas, Spyros and Mylonas, Nikos and Malounas, Ioannis and Rodias, Efthymios and Hellmann Santos, Christoph and Pekkeriet, Erik},
  journal={Sensors},
  volume={20},
  number={9},
  pages={2672},
  year={2020},
  publisher={MDPI}
}

@inproceedings{kruijff2012rescue,
  title={Rescue robots at earthquake-hit Mirandola, {Italy}: A field report},
  author={Kruijff, Geert-Jan M and Pirri, Fiora and Gianni, Mario and Papadakis, Panagiotis and Pizzoli, Matia and Sinha, Arnab and Tretyakov, Viatcheslav and Linder, Thorsten and Pianese, Emanuele and Corrao, Salvatore and others},
  booktitle={2012 IEEE international symposium on safety, security, and rescue robotics (SSRR)},
  pages={1--8},
  year={2012},
  organization={IEEE}
}

@article{conceptgraphs,
  author    = {Gu, Qiao and Kuwajerwala, Alihusein and Morin, Sacha and Jatavallabhula, {Krishna Murthy} and  Sen, Bipasha and Agarwal, Aditya and Rivera, Corban and Paul, William and Ellis, Kirsty and Chellappa, Rama and Gan, Chuang and {de Melo}, {Celso Miguel} and Tenenbaum, {Joshua B.} and Torralba, Antonio and Shkurti, Florian and Paull, Liam},
  title     = {ConceptGraphs: Open-Vocabulary 3D Scene Graphs for Perception and Planning},
  journal   = {International Conference on Robotics and Automation},
  year      = {2024},
}

@misc{dai2024optimalscenegraphplanning,
      title={Optimal Scene Graph Planning with Large Language Model Guidance}, 
      author={Zhirui Dai and Arash Asgharivaskasi and Thai Duong and Shusen Lin and Maria-Elizabeth Tzes and George Pappas and Nikolay Atanasov},
      year={2024},
      eprint={2309.09182},
      archivePrefix={arXiv},
      primaryClass={cs.RO},
      url={https://arxiv.org/abs/2309.09182}, 
}

@article{liu2024slideslam,
  title={Slideslam: Sparse, lightweight, decentralized metric-semantic slam for multirobot navigation},
  author={Liu, Xu and Lei, Jiuzhou and Prabhu, Ankit and Tao, Yuezhan and Spasojevic, Igor and Chaudhari, Pratik and Atanasov, Nikolay and Kumar, Vijay},
  journal={IEEE Transactions on Robotics},
  volume={41},
  pages={6529--6548},
  year={2025},
  publisher={IEEE}
}

@ARTICLE{fasterlio,  
author={Bai, Chunge and Xiao, Tao and Chen, Yajie and Wang, Haoqian and Zhang, Fang and Gao, Xiang},  
journal={IEEE Robotics and Automation Letters},   
title={Faster-LIO: Lightweight Tightly Coupled Lidar-Inertial Odometry Using Parallel Sparse Incremental Voxels},   
year={2022},  
volume={7},  
number={2},  
pages={4861-4868},  
doi={10.1109/LRA.2022.3152830}}

@article{liao2024clip,
  title={CLIP-GS: CLIP-Informed Gaussian Splatting for Real-time and View-consistent 3D Semantic Understanding},
  author={Liao, Guibiao and Li, Jiankun and Bao, Zhenyu and Ye, Xiaoqing and Wang, Jingdong and Li, Qing and Liu, Kanglin},
  journal={arXiv preprint arXiv:2404.14249},
  year={2024}
}

@inproceedings{radford2021learning,
  title={Learning transferable visual models from natural language supervision},
  author={Radford, Alec and Kim, Jong Wook and Hallacy, Chris and Ramesh, Aditya and Goh, Gabriel and Agarwal, Sandhini and Sastry, Girish and Askell, Amanda and Mishkin, Pamela and Clark, Jack and others},
  booktitle={International conference on machine learning},
  pages={8748--8763},
  year={2021},
  organization={PMLR}
}

@article{jatavallabhula2023conceptfusion,
  author    = {Jatavallabhula, {Krishna Murthy} and Kuwajerwala, Alihusein and Gu, Qiao and Omama, Mohd and Chen, Tao and Li, Shuang and Iyer, Ganesh and Saryazdi, Soroush and Keetha, Nikhil and Tewari, Ayush and Tenenbaum, {Joshua B.} and {de Melo}, {Celso Miguel} and Krishna, Madhava and Paull, Liam and Shkurti, Florian and Torralba, Antonio},
  title     = {ConceptFusion: Open-set Multimodal 3D Mapping},
  journal   = {Robotics: Science and Systems (RSS)},
  year      = {2023},
}

\end{document}